\documentclass[10pt,twocolumn,letterpaper]{article}

\usepackage[pagenumbers]{cvpr} 

\usepackage[dvipsnames]{xcolor}
\usepackage{multicol}
\usepackage{multirow}
\usepackage{graphicx}
\usepackage{amsmath}
\usepackage{amssymb}
\usepackage{dsfont}
\usepackage{pifont}         

\definecolor{cvprblue}{rgb}{0.21, 0.49, 0.74}
\definecolor{scheme_red}{RGB}{216, 0, 115}
\definecolor{scheme_blue}{RGB}{27, 161, 226}
\newcommand{\cmark}{\ding{51}}

\usepackage[pagebackref,breaklinks,colorlinks,citecolor=cvprblue]{hyperref}

\def\paperID{5601} 
\def\confName{CVPR}
\def\confYear{2024}

\title{OneFormer3D: One Transformer for Unified Point Cloud Segmentation}

\author{
Maxim Kolodiazhnyi, Anna Vorontsova, Anton Konushin, Danila Rukhovich\\
Samsung Research \\
{\tt \small \{m.kolodiazhn,\ a.vorontsova,\ a.konushin,\ d.rukhovich\}@samsung.com}}

\begin{document}
\maketitle
\begin{abstract}
Semantic, instance, and panoptic segmentation of 3D point clouds have been addressed using task-specific models of distinct design. Thereby, the similarity of all segmentation tasks and the implicit relationship between them have not been utilized effectively. This paper presents a unified, simple, and effective model addressing all these tasks jointly. The model, named OneFormer3D, performs instance and semantic segmentation consistently, using a group of learnable kernels, where each kernel is responsible for generating a mask for either an instance or a semantic category. These kernels are trained with a transformer-based decoder with unified instance and semantic queries passed as an input. Such a design enables training a model end-to-end in a single run, so that it achieves top performance on all three segmentation tasks simultaneously. Specifically, our OneFormer3D ranks \textbf{1\textsuperscript{st}} and sets a new state-of-the-art (\textbf{+2.1} mAP\textsubscript{50}) in the ScanNet test leaderboard. We also demonstrate the state-of-the-art results in semantic, instance, and panoptic segmentation of ScanNet (+21 PQ), ScanNet200 (+3.8 mAP\textsubscript{50}), and S3DIS (+0.8 mIoU) datasets.
\end{abstract}

\begin{figure}[t!]
    \centering
    \includegraphics[width=0.82\linewidth]{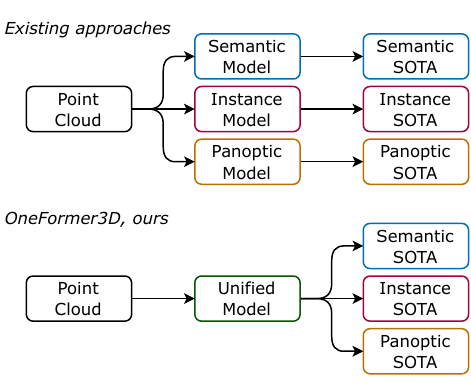}
    \caption{Traditional 3D point cloud segmentation methods address different tasks with task-specific models to achieve the best performance. We propose OneFormer3D, a 3D segmentation framework that tackles semantic, instance, and panoptic segmentation tasks with a multi-task train-once design.}
    \label{fig:teaser}
\end{figure}

\section{Introduction}

3D point cloud segmentation is the task of grouping points into meaningful segments. Such a segment may comprise points of the same semantic category or belonging to the same single object (an instance). Semantic-based and instance-based grouping give rise to three formulations of the segmentation task: semantic, instance, and panoptic. Semantic segmentation outputs a mask for each semantic category, so that each point in a point cloud gets assigned with a semantic label. Instance segmentation returns a set of masks of individual objects; since some regions cannot be treated as an distinguishable object but rather serve as a background (like a \textit{floor} or a \textit{ceiling}), only a part of points in a point cloud is being labeled. Panoptic segmentation is the most general formulation: it implies predicting a mask for each foreground object (\textit{thing}), and a semantic label for each background point (\textit{stuff}).

Despite all three 3D segmentation tasks actually imply predicting a set of masks, they are typically solved with models of completely different architectures. 3D semantic segmentation methods rely on U-Net-like networks~\cite{qi2017pointnet++, qian2022pointnext, thomas2019kpconv, lin2023pointmetabase, zhao2021pointtransformer, wu2022pointtransformerv2, choy2019minkowski}. 3D instance segmentation methods combine semantic segmentation models with aggregation schemes based either on clustering~\cite{vu2022softgroup, he2021dyco3d, chen2021hais, liang2021sstnet, jiang2020pointgroup, zhao2023pbnet}, object detection~\cite{hou20193dsis, kolodiazhnyi2023td3d}, or transformer decoders~\cite{schult2023mask3d, sun2023spformer}. 3D panoptic segmentation methods~\cite{wu2021scenegraphfusion, narita2019panopticfusion, yang2021tuppermap} perform panoptic segmentation in 2D images, than lift the predicted masks into 3D space and aggregate them point-wise. The question naturally arises: is it possible to tackle all three 3D segmentation tasks jointly with a single unified approach? 

Recenty, various ways of unifying 2D segmentation methods have been proposed~\cite{wu2022dknet, cheng2021maskformer, cheng2022mask2former}. All these methods train a single panoptic model on all three tasks, so that high performance is obtained without changing the network architecture. Still, the best performance is achieved when the model is trained for each task separately. As can be expected, such a training policy results in three times larger time- and memory footprint: training lasts longer and produces different sets of model weights for each task. This drawback was eliminated in a recent OneFormer \cite{jain2023oneformer} -- a multi-task unified image segmentation approach, which outperforms existing state-of-the-arts in all three image segmentation tasks after training on a panoptic dataset in a joint fashion.

Following the same path, we propose OneFormer3D, the first multi-task unified 3D segmentation framework (Fig. \ref{fig:teaser}). Using a well-known SPFormer \cite{sun2023spformer} baseline, we add semantic queries in parallel with instance queries in a transformer decoder to unify predicting semantic and instance segmentation masks. Then, we identify the reasons for unstable performance of transformer-based 3D instance segmentation, and resolve the issues with a novel query selection mechanism and a new efficient matching strategy. Finally, we come up with a single unified model trained only once, that outperforms 3D semantic, 3D instance, and 3D panoptic segmentation methods -- even though they are specifically tuned for each task.

To summarize, our contributions are as follows:
\begin{itemize}
\item OneFormer3D -- the first multi-task unified 3D segmentation framework, which allows training a single model on a common panoptic dataset to solve three segmentation tasks jointly;
\item A novel query selection strategy and an efficient matching strategy without Hungarian algorithm, that should be used in combination for the best quality;
\item State-of-the-art results in 3D semantic, 3D instance, and 3D panoptic segmentation in three indoor benchmarks: ScanNet~\cite{dai2017scannet}, ScanNet200~\cite{rozenberszki2022lground}, and S3DIS~\cite{armeni2016s3dis}.
\end{itemize}

\section{Related Work}

\subsection{3D Point Cloud Segmentation}

\paragraph{3D Semantic Segmentation.} Learning-based methods for semantic segmentation of 3D point clouds leverage U-Net-like models to process either 3D points (point-based) or voxels (voxel-based). Point-based methods exploit hand-crafted aggregation mechanisms \cite{qi2017pointnet++, qian2022pointnext, thomas2019kpconv, lin2023pointmetabase} or transformer blocks \cite{zhao2021pointtransformer, wu2022pointtransformerv2} for direct processing of points. Voxel-based methods transform a point cloud of an irregular structure to a regular voxel grid, and pass these voxels through dense \cite{hou20193dsis} or sparse \cite{choy2019minkowski} 3D convolutional network. Considering time- and memory efficiency, we opt for a sparse convolutional U-Net as a backbone, and combine it with a transformer decoder; to the best of our knowledge, OneFormer3D is the ever-first method using such a decoder to solve the 3D semantic segmentation task.

\paragraph{3D Instance Segmentation.} Instance segmentation of 3D point clouds is typically addressed with 3D semantic segmentation followed by per-point features aggregation. Earlier approaches can be classified into top-down proposal-based methods~\cite{kolodiazhnyi2023td3d, hou20193dsis, sun2023neuralbf, yi2019gspn} or bottom-up grouping-based methods~\cite{vu2022softgroup, he2021dyco3d, chen2021hais, liang2021sstnet, jiang2020pointgroup}. Current state-of-the-art results belong to recently emerged transformer-based methods, that outperform the predecessors in both accuracy \cite{schult2023mask3d} and inference speed \cite{sun2023spformer}. We consider SPFormer \cite{sun2023spformer} as our baseline, and extend it, so that it solves not a single 3D instance segmentation but all three 3D segmentation tasks.

\paragraph{3D Panoptic Segmentation.} Panoptic segmentation of 3D point clouds is an underexplored problem, with only  few existing solutions~\cite{wu2021scenegraphfusion, narita2019panopticfusion, yang2021tuppermap}; all of them being trained and validated only on the ScanNet dataset. These methods apply panoptic segmentation to a set of RGB images, lift the predicted 2D panoptic masks into 3D space, and obtain final 3D panoptic masks through aggregation. On the contrary, our OneFormer3D does not require additional RGB data to achieve state-of-the-art panoptic segmentation quality.

\subsection{Unified 2D Image Segmentation}
 
Unified 2D segmentation has been extensively researched over the past few years, resulting in a variety of methods proposed~\cite{wu2022dknet, cheng2021maskformer, cheng2022mask2former}. K-Net \cite{zhang2021knet} uses a convolutional network with dynamic learnable instance and semantic kernels with bipartite matching. MaskFormer \cite{cheng2021maskformer} is a transformer-based architecture for mask classification. It was inspired by object detection~\cite{carion2020detr}, where the image is first fed to the encoder to obtain queries, then the decoder outputs proposals based on these queries. Mask2Former \cite{cheng2022mask2former} extends MaskFormer with learnable queries, deformable multi-scale attention in the decoder, and a masked cross-attention, setting a new state-of-the-art in all three segmentation tasks. However, all methods mentioned above still require training the model individually for each task to achieve the best performance. OneFormer \cite{jain2023oneformer} was the pioneer 2D image segmentation approach, that employs task-conditioned joint training strategy and achieves state-of-the-art results in three segmentation tasks simultaneously with a single model. In a similar fashion, we build OneFormer3D for 3D point cloud segmentation.

\section{Proposed Method}

\begin{figure*}
    \centering
    \includegraphics[width=0.85\linewidth]{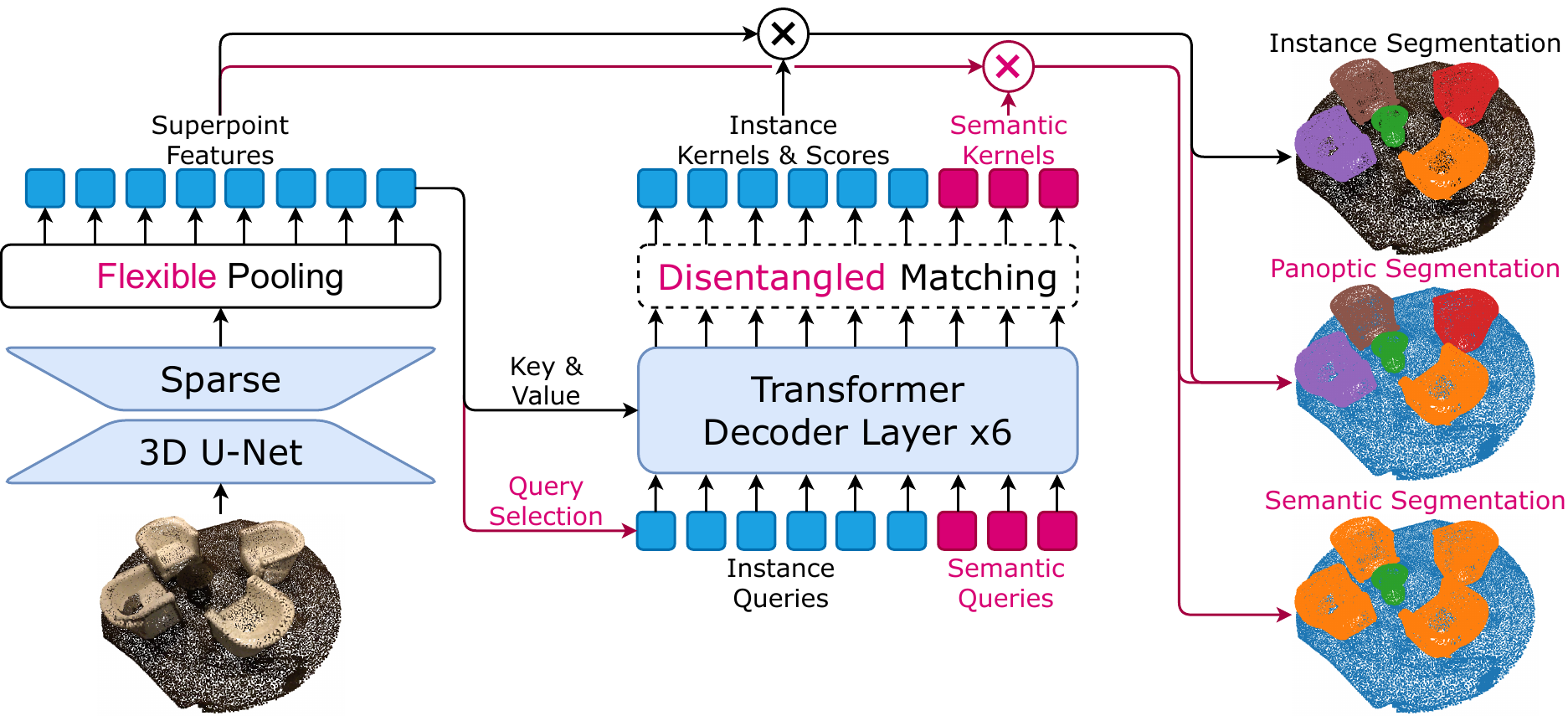}
    \caption{The OneFormer3D framework is based on SPFormer (\textcolor{scheme_blue}{blue}), but features a number of improvements (\textcolor{scheme_red}{red}). Once obtained a 3D point cloud as as input, our trained model solves 3D instance, 3D semantic, and 3D panoptic segmentation tasks. The dotted line depicts components that are applied only during the training.}
    \label{fig:scheme}
\end{figure*}

The general scheme of OneFormer3D is shown in Fig. \ref{fig:scheme}, with a baseline components depicted in blue and novelty points highlighted with a red color. Our framework is inherited from SPFormer \cite{sun2023spformer}, which was originally proposed to tackle 3D instance segmentation. SPFormer is chosen due to its straightforward pipeline, fast inference, and small memory footprint during both training and inference; yet, any modern 3D instance segmentation method with a transformer decoder can be used instead (e.g., Mask3D \cite{schult2023mask3d}).

First, a sparse 3D U-net extracts point-wise features (Sec. \ref{sec:backbone}). Then, these features pass through a flexible pooling, that obtains superpoint features through simply averaging features of points in a superpoint. 
Superpoint features serve as keys and values for a transformer decoder (Sec. \ref{sec:decoder}), that also accepts learnable semantic and instance queries as inputs. The decoder captures superpoints information via a cross-attention mechanism, and outputs a set of learned kernels, each representing a single object mask of an instance identity (from an instance query) or a semantic region (from a semantic query. A disentangled matching strategy is adopted to train instance kernels in an end-to-end manner (Sec. \ref{sec:training}). As a result, a trained OneFormer3D can seamlessly solve semantic, instance, and panoptic segmentation (Sec. \ref{sec:inference}).

\subsection{Backbone and Pooling}
\label{sec:backbone}
\paragraph{Sparse 3D U-Net.} Assuming that an input point cloud contains $N$ points, the input can be formulated as $\boldsymbol{P} \in \mathbb{R}^{N \times 6}$. Each 3D point is parameterized with three colors \textit{r}, \textit{g}, \textit{b}, and three coordinates \textit{x}, \textit{y}, \textit{z}. Following \cite{choy2019minkowski}, we voxelize point cloud, and use a U-Net-like backbone composed of sparse 3D convolutions to extract point-wise features $\boldsymbol{P}' \in \mathbb{R}^{N \times C}$.

\paragraph{Flexible pooling.} For a greater flexibility, we implement pooling based on either superpoints or voxels. In a superpoint pooling scenario, superpoint features $\mathbf S \in \mathbb{R} {^{M \times C}}$ are obtained via average pooling of point-wise features $\mathbf P' \in \mathbb{R} {^{N \times C}}$ w.r.t. pre-computed superpoints \cite{landrieu2018superpoint}. Without loss of generality, we suppose that there are $M$ superpoints in an input point cloud. In a voxel pooling scenario, we pool backbone features w.r.t. voxel grid. Voxelization is a trivial operation with a negligible computational overhead; accordingly, it can be preferred to computationally-heavy superpoint clustering in resource-constrained usage scenarios. We refer to this superpoint-based / voxel-based pooling as \textit{flexible pooling}. This procedure transforms an input point cloud comprised of millions of points into only hundreds of superpoints or thousands of voxels, which significantly reduces the computational cost of subsequent processing.

\subsection{Query Decoder}
\label{sec:decoder}

A query decoder takes $K_{ins}+K_{sem}$ queries as inputs and transforms them into $K_{ins}+K_{sem}$ kernels. Then, superpoint features are convolved with these kernels to produce $K_{ins}$ instance and $K_{sem}$ semantic masks, respectively. The architecture of a query decoder is inherited from SPFormer~\cite{sun2023spformer}: similarly, six sequential transformer decoder layers employ self-attention on queries and cross-attention with keys and values from superpoint features. Semantic queries are initialized randomly, same as in existing 3D instance segmentation methods~\cite{sun2023spformer, schult2023mask3d}. Instance queries are initialized through the \textit{query selection} strategy.

\paragraph{Query selection.}
State-of-the-art 2D object detection and 2D instance segmentation methods~\cite{zhu2020deformabledetr, zhang2022dino, li2023maskdino} initialize queries using advanced strategies, usually referred to as \textit{query selection}. Specifically, input queries are initialized with features from a transformer encoder, sampled based on an objectness score. This score is estimated by the same model, which is guided by an additional objectness loss during the training. The described technique is proved to speed up the training, while jointly improving the overall accuracy. Yet, to the best of our knowledge, a similar approach was never applied in 3D object detection or 3D segmentation. So, we aim to close this gap with a simplified version of query selection adapted for 3D data and a non-transformer encoder. Particularly, we initialize queries with backbone features after a flexible pooling. By a query selection, we randomly select only a half of initialized queries for an extra augmentation during the training. During the inference, we initialize queries similarly, but do not filter queries to keep all input information. 

\subsection{Training}
\label{sec:training}

To train a transformer-based method end-to-end, we need to define a cost function between queries and ground truth objects, develop a matching strategy that minimizes this cost function, and formulate a loss function being applied to the matched pairs.

\paragraph{Cost function.}
Following SPFormer \cite{sun2023spformer}, we use a pairwise matching cost $\mathcal{C}_{ik}$ to measure the similarity of the $i$-th proposal and the $k$-th ground truth. $\mathcal{C}_{ik}$ is derived from a classification probability and a superpoint mask matching cost $\mathcal{C}^{mask}_{ik}$:
\begin{equation}
    \label{equ:C}
    \mathcal{C}_{ik} = -\lambda \cdot  p_{i,c_k}+\mathcal{C}^{mask}_{ik},
\end{equation}
where $p_{i,c_k}$ indicates the probability of $i$-th proposal belonging to the $c_k$ semantic category. In our experiments, we use $\lambda_{cls} = 0.5$. The superpoint mask matching cost $\mathcal{C}^{mask}_{ik}$ is a sum of a binary cross-entropy (BCE) and a Dice loss with a Laplace smoothing:
\begin{equation}
    \label{equ:Cmask}
    \mathcal{C}^{mask}_{ik} = \text{BCE}(m_i,m^{gt}_k)+1-2\frac{ m_i\cdot m^{gt}_k+1}{\left | m_i \right | +\left | m^{gt}_k \right |+1},
\end{equation}
where $m_i$ and $m^{gt}_k$ are a predicted and ground truth mask of a superpoint, respectively.

\paragraph{Disentangled matching.}

Previous state-of-the-art 2D transformer-based methods~\cite{carion2020detr, li2023maskdino, cheng2021maskformer, cheng2022mask2former} and 3D transformer-based methods~\cite{schult2023mask3d, sun2023spformer} exploit a bipartite matching strategy based on a Hungarian algorithm \cite{kuhn1955hungarian}. This commonly-used approach has though a major drawback: an excessive number of meaningful matches between proposals and ground truth instances makes the training process long-lasting and unstable. 

On the contrary, we perform a simple trick that eliminates the need for resource-exhaustive Hungarian matching. Since an instance query is initialized with features of a superpoint, this instance query can be unambiguously matched with this superpoint. We assume that a superpoint can belong only to one instance, that gives a correspondence between a superpoint and a ground truth object. By bringing everything together, we can establish the correspondence between a ground truth object, a superpoint, an instance query, and an instance proposal derived from this instance query. Finally, by skipping intermediate correspondences, we can directly match an instance proposal to a ground truth instance. The obtained correspondence \textit{disentangles} the bipartite graph of proposals and ground truth instances, that is why we refer to it as our \textit{disentangled matching}.

Still, the number of proposals exceeds the number of ground truth instances, so we need to filter out proposals that do not correspond to ground truth objects to obtain a bipartite matching. The disentangled matching trick simplifies cost function optimization, as we can set the most weights in a cost matrix to infinity: 
\begin{equation}
\mathcal{\hat{C}}_{ik}=\left\{  
    \begin{array}{rl} 
    \mathcal{C}_{ik} & \text{if $i$-th superpoint $\in$ $k$-th object} \\
    +\infty & \text{otherwise}
    \end{array}
\right.
\end{equation}
In a standard scenario, all values of a cost matrix are non-infinite. Accordingly, the optimal solution can be obtained via a Hungarian matching with a computational complexity of $O(K_{ins}^3)$. Our disentangled matching is though notably more efficient, having a $O(K_{ins})$ complexity. For a ground truth instance $k$, we only need to select the proposal $i$ with the least $\mathcal{\hat{C}}_{ik}$. Since there is only one non-infinite value per proposal, this operation is trivial and can be performed in a linear time.


\paragraph{Loss.}

After matching proposals with ground truth instances, instance losses can finally be calculated. Classification errors are penalized with a cross-entropy loss $\mathcal{L}_{cls}$. Besides, for each match between a proposal and a ground truth instance, we compute the superpoint mask loss as a sum of binary cross-entropy $\mathcal{L}_{bce}$ and a Dice loss $\mathcal{L}_{dice}$.

$K_{sem}$ semantic queries correspond to ground truth masks of $K_{sem}$ semantic categories given in a fixed order, so no specific matching is required. The semantic loss $L_{sem}$ is defined as a binary cross-entropy.

The total loss $\mathcal{L}$ is formulated as:
\begin{equation}
\label{equ:L}
\mathcal{L}=\beta \cdot \mathcal{L}_{cls}+\mathcal{L}_{bce}+\mathcal{L}_{dice} + L_{sem},
\end{equation}
where $\beta = 0.5$ as in \cite{sun2023spformer}.

\subsection{Inference}
\label{sec:inference}
During inference, given an input point cloud, OneFormer3D directly predicts $K_{sem}$ semantic masks and $K_{ins}$ instance with classification scores $p_i$, $i \in {1,\ ...\ K_{ins}}$, where each mask $m_i$ is a set of superpoints. Then, we convolve superpoint features $\mathbf S \in \mathbb{R} {^{M \times C}}$ with each predicted kernel $l_i \in \mathbb{R} {^{1 \times C}}$ to get a mask $m_i \in \mathbb{R} {^{M \times 1}}$: $m_i = \mathbf S * l_i$. The final binary segmentation masks are obtained by thresholding probability scores. Besides, for $m_i$, we calculate a mask score $q_i \in [0, 1]$ by averaging probabilities exceeding the threshold, and use it to set an initial ranking score $s_i$: $s_i = p_i \cdot q_i$. Finally, $s_i$ values are leveraged for re-ranking predicted instances using matrix-NMS~\cite{wang2020solov2}.

Panoptic prediction is obtained from instance and semantic outputs. It is initialized with estimated semantics, then, instance predictions are overlaid consequently, sorted by a ranking score in an increasing order.

\section{Experiments}

\subsection{Experimental Settings}

\paragraph{Datasets.} The experiments are conducted on ScanNet~\cite{dai2017scannet}, ScanNet200~\cite{rozenberszki2022lground}, and S3DIS~\cite{armeni2016s3dis} datasets.
ScanNet~\cite{dai2017scannet} contains 1613 scans divided into training, validation, and testing splits of 1201, 312, and 100 scans, respectively. 3D instance segmentation is typically evaluated using 18 object categories. Two more categories (\textit{wall} and \textit{floor}) are added for semantic and panoptic evaluation. We report results on both validation and hidden test splits. ScanNet200~\cite{rozenberszki2022lground} extends the original ScanNet semantic annotation with fine-grained categories with the long-tail distribution, resulting in 198 instance with 2 more semantic classes. The training, validation, and testing splits are similar to the original ScanNet dataset. The S3DIS dataset~\cite{armeni2016s3dis} features 272 scenes within 6 large areas. Following the standard evaluation protocol, we assess the segmentation quality on scans from Area-5, and via 6 cross-fold validation, using 13 semantic categories in both settings. Following the official \cite{armeni2016s3dis} split, we classify these 13 categories as either structured or furniture, and define 5 furniture categories (\textit{table}, \textit{chair}, \textit{sofa}, \textit{bookcase}, and \textit{board}) as \textit{thing}, and the remaining eight categories as \textit{stuff} for panoptic evaluation. 

\paragraph{Metrics.} 
We use mIoU to measure the quality of 3D semantic segmentation. For 3D instance segmentation, we report a mean average precision (mAP), which is an average of scores obtained with IoU thresholds set from 50\% to 95\%, with a step size of 5\%. mAP\textsubscript{50} and mAP\textsubscript{25} denote the scores with IoU thresholds of 50\% and 25\%, respectively. Additionally, we calculate mean precision (mPrec), and mean recall (mRec) for S3DIS, following the standard evaluation protocol established in this benchmark. The accuracy of panoptic predictions is assessed with the PQ score \cite{kirillov2019panoptic}; we also report PQ\textsubscript{th} and PQ\textsubscript{st}, estimated for \textit{thing} and \textit{stuff} categories, respectively. 

\paragraph{Implementation Details.}
Our OneFormer3D is implemented in MMDetection3D framework \cite{mmdet3d2020}. All training details are inherited from SPFormer \cite{sun2023spformer}, including using AdamW optimizer with an initial learning rate of 0.0001, weight decay of 0.05, batch size of 4, and polynomial scheduler with a base of 0.9 for 512 epochs. We apply the standard augmentations: horizontal flipping, random rotations around the z-axis, elastic distortion, and random scaling. On ScanNet and ScanNet200, we apply graph-based superpoint clusterization \cite{landrieu2018superpoint} and use a voxel size of 2cm. On S3DIS, voxel size is set to 5cm due to larger scenes.

\subsection{Comparison to Prior Work}

We compare our OneFormer3D with previous art on three indoor benchmarks: ScanNet \cite{dai2017scannet}, S3DIS \cite{armeni2016s3dis}, and ScanNet200 \cite{rozenberszki2022lground} in Tab. \ref{tab:scannet}, \ref{tab:s3dis}, and \ref{tab:scannet200}, respectively. On the ScanNet validation split, we set a new state-of-the art in instance, semantic, and panoptic segmentation tasks with a unified approach. Specifically, the instance segmentation scores increase by +2.9 mAP\textsubscript{25}, +4.4 mAP\textsubscript{25}, and +4.1 mAP compared to SPFormer~\cite{sun2023spformer} and a more recent Mask3D \cite{schult2023mask3d}, which is a notable improvement. Besides, OneFormer3D scores top-1 in the ScanNet hidden test leaderboard at 17 Nov. 2023 with 80.1 mAP\textsubscript{50} (+2.1 w.r.t. Mask3D), and incredible 89.6 mAP\textsubscript{25} (+2.1 w.r.t. TD3D \cite{kolodiazhnyi2023td3d}). At the same time, OneFormer3D supersedes PointTransformerV2\cite{wu2022pointtransformerv2}, a state-of-the-art semantic segmentation method, by +1.2 mIoU. Panoptic segmentation has not been investigated so extensively as the other two segmentation tasks, so this track is represented with few baselines demonstrating mediocre performance. Respectively, the improvement here is especially tangible: OneFormer3D outperforms TUPPer-Map by +21.0 PQ, hitting as high as 71.2.

\begin{table}[htb!]
\centering
\small
\begin{tabular}{llcc}
\toprule
\multirow{2}{*}{Method} & \multirow{2}{*}{Presented at} & Box & Box \\
& & mAP\textsubscript{25} & mAP\textsubscript{50} \\
\midrule
VoteNet\cite{qi2019votenet} & ICCV'19 & 58.6 & 33.5 \\
PBNet\cite{zhao2023pbnet} & ICCV'23 & 69.3 & 60.1 \\
FCAF3D\cite{rukhovich2022fcaf3d} & ECCV'22 & 71.5 & 57.3 \\
SoftGroup\cite{vu2022softgroup} & CVPR'22 & 71.6 & 59.4 \\
TR3D\cite{rukhovich2023tr3d} & ICIP'23 & 72.9 & 59.3 \\
CAGroup3D\cite{wang2022cagroup3d} & NeurIPS'22 & 75.1 & 61.3 \\
\multicolumn{2}{l}{\textbf{OneFormer3D}} & \textbf{76.9} & \textbf{65.3} \\
\bottomrule
\end{tabular}
\caption{Comparison of existing 3D object detection methods on the ScanNet validation split.}
\label{tab:detection}
\end{table}

\begin{table*}
\centering
\small
\begin{tabular}{lllccccccc}
\toprule
& \multirow{2}{*}{Method} & \multirow{2}{*}{Presented at} & \multicolumn{3}{c}{Instance} & Semantic & \multicolumn{3}{c}{Panoptic} \\
& & & mAP\textsubscript{25} & mAP\textsubscript{50} & mAP & mIoU & PQ & PQ\textsubscript{th} & PQ\textsubscript{st} \\
\midrule
\multicolumn{2}{l}{\textit{Validation split}} & & & & & & & & \\
& 3D-SIS\cite{hou20193dsis} & CVPR'19 & 35.7 & 18.7 & & & & & \\
& GSPN\cite{yi2019gspn} & CVPR'19 & 53.4 & 37.8 & 19.3 & & & & \\
& NeuralBF\cite{sun2023neuralbf} & WACV'23 & 71.1 & 55.5 & 36.0 & & & & \\
& PointGroup\cite{jiang2020pointgroup} & CVPR'20 & 71.3 & 56.7 & 34.8 & & & & \\
& OccuSeg\cite{han2020occuseg} & CVPR'20 & 71.9 & 60.7 & 44.2 & & & & \\
& DyCo3D\cite{he2021dyco3d} & CVPR'21 & 72.9 & 57.6 & 35.4 & & & & \\
& SSTNet\cite{liang2021sstnet} & ICCV'21 & 74.0 & 64.3 & 49.4 & & & & \\
& HAIS\cite{chen2021hais} & ICCV'21 & 75.6 & 64.4 & 43.5 & & & & \\
& DKNet\cite{wu2022dknet} & ICCV'22 & 76.9 & 66.7 & 50.8 & & & & \\
& SoftGroup\cite{vu2022softgroup} & CVPR'22 & 78.9 & 67.6 & 45.8 & & & & \\
& PBNet\cite{zhao2023pbnet} & ICCV'23 & 78.9 & 70.5 & 54.3 & & & & \\
& TD3D\cite{kolodiazhnyi2023td3d} & WACV'24 & 81.9 & 71.2 & 47.3 & & & & \\
& ISBNet\cite{ngo2023isbnet} & CVPR'23 & 82.5 & 73.1 & 54.5 & & & & \\
& SPFormer\cite{sun2023spformer} & AAAI'23 & 82.9 & 73.9 & 56.3 & & & & \\
& Mask3D\cite{schult2023mask3d} & ICRA'23 & 83.5 & 73.7 & 55.2 & & & & \\
& PointNet++\cite{qi2017pointnet++} & NeurIPS'17 & & & & 53.5 & & & \\
& PointConv\cite{wu2019pointconv} & CVPR'19 & & & & 61.0 & & & \\
& PointASNL\cite{yan2020pointasnl} & CVPR'20 & & & & 63.5 & & & \\
& KPConv\cite{thomas2019kpconv} & ICCV'19 & & & & 69.2 & & & \\
& PointTransformer\cite{zhao2021pointtransformer} & ICCV'21 & & & & 70.6 & & & \\
& PointNeXt-XL\cite{qian2022pointnext} & NeurIPS'22 & & & & 71.5 & & & \\
& MinkUNet\cite{choy2019minkowski} & CVPR'19 & & & & 72.2 & & & \\
& PointMetaBase-XXL\cite{lin2023pointmetabase} & CVPR'23 & & & & 72.8 & & & \\
& PointTransformerV2\cite{wu2022pointtransformerv2} & NeurIPS'22 & & & & 75.4 & & & \\
& SceneGraphFusion\cite{wu2021scenegraphfusion} & CVPR'21 & & & & & 31.5 & 30.2 & 43.4 \\
& PanopticFusion\cite{narita2019panopticfusion} & IROS'19 & & & & & 33.5 & 30.8 & 58.4 \\
& TUPPer-Map\cite{yang2021tuppermap} & IROS'21 & & & & & 50.2 & 47.8 & 71.5 \\
& \textbf{OneFormer3D (ours)} & & \textbf{86.4} & \textbf{78.1} & \textbf{59.3} & \textbf{76.6} & \textbf{71.2} & \textbf{69.6} & \textbf{86.1} \\
\midrule
\multicolumn{2}{l}{\textit{Hidden test split at 17 Nov. 2023}} & & & & & & & & \\
& NeuralBF\cite{sun2023neuralbf} & WACV'23 & 71.8 & 55.5 & 35.3 & & & & \\
& DyCo3D\cite{he2021dyco3d} & CVPR'21 & 76.1 & 64.1 & 39.5 & & & & \\
& PointGroup\cite{jiang2020pointgroup} & CVPR'20 & 77.8 & 63.6 & 40.7 & & & & \\
& SSTNet\cite{liang2021sstnet} & ICCV'21 & 78.9 & 69.8 & 50.6 & & & & \\
& HAIS\cite{chen2021hais} & ICCV'21 & 80.3 & 69.9 & 45.7 & & & & \\
& DKNet\cite{wu2022dknet} & ICCV'22 & 81.5 & 71.8 & 53.2 & & & & \\
& ISBNet\cite{ngo2023isbnet} & CVPR'23 & 83.5 & 75.7 & 55.9 & & & & \\
& SPFormer\cite{sun2023spformer} & AAAI'23 & 85.1 & 77.0 & 54.9 & & & & \\
& SoftGroup\cite{vu2022softgroup} & CVPR'22 & 86.5 & 76.1 & 50.4 & & & & \\
& Mask3D\cite{schult2023mask3d} & ICRA'23 & 87.0 & 78.0 & 56.6 & & & & \\
& TD3D\cite{kolodiazhnyi2023td3d} & WACV'24 & 87.5 & 75.1 & 48.9 & & & & \\
& \textbf{OneFormer3D (ours)} & & \textbf{89.6} & \textbf{80.1} & \textbf{56.6} & & & & \\
\bottomrule
\end{tabular}
\caption{Comparison of the existing segmentation methods on ScanNet. Our OneFormer3D sets the new state-of-the art in all segmentation tasks: instance, semantic, and panoptic.}
\label{tab:scannet}
\end{table*}

Besides, we adopt our OneFormer3D to 3D object detection by enclosing predicted 3D instances with tight axis-aligned 3D bounding boxes. The comparison with existing 3D object detection methods in presented in Tab. \ref{tab:detection}. As can be seen, OneFormer3D achieves +4.0 mAP\textsubscript{50} w.r.t. a strong CAGroup3D\cite{wang2022cagroup3d} baseline, hence setting a new state-of-the-art in 3D object detection with 65.1 mAP\textsubscript{50} with no extra training.

\begin{table*}
\centering
\small
\begin{tabular}{llcccccccc}
\toprule
& \multirow{2}{*}{Method} & \multicolumn{4}{c}{Instance} & Semantic & \multicolumn{3}{c}{Panoptic} \\
& & mAP\textsubscript{50} & mAP & mPrec\textsubscript{50} & mRec\textsubscript{50} & mIoU & PQ & PQ\textsubscript{th} & PQ\textsubscript{st} \\
\midrule
\multicolumn{2}{l}{\textit{Area-5 validation}} & & & & & & & & \\
& PointGroup\cite{jiang2020pointgroup} & 57.8 & & 61.9 & 62.1 & & & & \\
& DyCo3D\cite{he2021dyco3d} & & & 64.3 & 64.2 & & & & \\
& SSTNet\cite{liang2021sstnet} & 59.3 & 42.7 & 65.5 & 64.2 & & & & \\
& DKNet\cite{wu2022dknet} & & & 70.8 & 65.3 & & & & \\
& HAIS\cite{chen2021hais} & & & 71.1 & 65.0 & & & & \\
& TD3D\cite{kolodiazhnyi2023td3d} & 65.1 & 48.6 & 74.4 & 64.8 & & & & \\
& SoftGroup\cite{vu2022softgroup} & 66.1 & 51.6 & 73.6 & 66.6 & & & & \\
& PBNet\cite{zhao2023pbnet} & 66.4 & 53.5 & 74.9 & 65.4 & & & & \\
& SPFormer\cite{sun2023spformer} & 66.8 & & 72.8 & 67.1 & & & & \\
& Mask3D\cite{schult2023mask3d} & 71.9 & 57.8 & 74.3 & 63.7 & & & & \\
& SegGCN\cite{lei2020seggcn} & & & & & 63.6 & & & \\
& MinkUNet\cite{choy2019minkowski} & & & & & 65.4 & & & \\
& PAConv\cite{xu2021paconv} & & & & & 66.6 & & & \\
& KPConv\cite{thomas2019kpconv} & & & & & 67.1 & & & \\
& PointTransformer\cite{zhao2021pointtransformer} & & & & & 70.4 & & & \\
& PointNeXt-XL\cite{qian2022pointnext} & & & & & 70.5 & & & \\
& PointTransformerV2\cite{wu2022pointtransformerv2} & & & & & 71.6 & & & \\
& \textbf{OneFormer3D (ours)} & \textbf{72.0} & \textbf{58.7} & \textbf{79.7} & \textbf{73.0} & \textbf{72.4} & \textbf{62.2} & \textbf{58.4} & \textbf{65.5} \\
\midrule
\multicolumn{2}{l}{\textit{6-fold cross-validation}} & & & & & & & & \\
& PointGroup\cite{jiang2020pointgroup} & 64.0 & & 69.6 & 69.2 & & & & \\
& HAIS\cite{chen2021hais} & & & 73.2 & 69.4 & & & & \\
& SSTNet\cite{liang2021sstnet} & 67.8 & 54.1 & 73.5 & 73.4 & & & & \\
& DKNet\cite{wu2022dknet} & & & 75.3 & 71.1 & & & & \\
& TD3D\cite{kolodiazhnyi2023td3d} & 68.2 & 56.2 & 76.3 & 74.0 & & & & \\
& SoftGroup\cite{vu2022softgroup} & 68.9 & 54.4 & 75.3 & 69.8 & & & & \\
& SPFormer\cite{sun2023spformer} & 69.2 & & 74.0 & 71.1 & & & & \\
& PBNet\cite{zhao2023pbnet} & 70.6 & 59.5 & 80.1 & 72.9 & & & & \\
& Mask3D\cite{schult2023mask3d} & 74.3 & 61.8 & 76.5 & 66.2 & & & & \\
& PointNet++\cite{qi2017pointnet++} & & & & & 56.7 & & & \\
& MinkUNet\cite{choy2019minkowski} & & & & & 69.1 & & & \\
& KPConv\cite{thomas2019kpconv} & & & & & 70.6 & & & \\
& PointTransformer\cite{zhao2021pointtransformer} & & & & & 73.5 & & & \\
& PointNeXt-XL\cite{qian2022pointnext} & & & & & 74.9 & & & \\
& \textbf{OneFormer3D (ours)} & \textbf{75.8} & \textbf{63.0} & \textbf{82.3} & \textbf{74.1} & \textbf{75.0} & \textbf{68.5} & \textbf{61.5} & \textbf{74.5} \\
\bottomrule
\end{tabular}
\caption{Comparison of existing segmentation methods on S3DIS. Our OneFormer3D sets the new state-of-the art in all segmentation tasks: instance, semantic, and panoptic.}
\label{tab:s3dis}
\end{table*}

\begin{table*}
\centering
\small
\begin{tabular}{lccccccc}
\toprule
\multirow{2}{*}{Method} & \multicolumn{3}{c}{Instance} & Semantic & \multicolumn{3}{c}{Panoptic} \\
& mAP\textsubscript{25} & mAP\textsubscript{50} & mAP & mIoU & PQ & PQ\textsubscript{th} & PQ\textsubscript{st}\\
\midrule
PointGroup\cite{jiang2020pointgroup} & & 24.5 & & & & & \\
PointGroup + LGround\cite{rozenberszki2022lground} & & 26.1 & & & & & \\
TD3D\cite{kolodiazhnyi2023td3d} & 40.4 & 34.8 & 23.1 & & & & \\
Mask3D\cite{schult2023mask3d} & 42.3 & 37.0 & 27.4 & & & & \\
MinkUNet\cite{choy2019minkowski} & & & & 25.0 & & & \\
MinkUNet + LGround\cite{rozenberszki2022lground} & & & & 28.9 & & & \\
\textbf{OneFormer3D (ours)} & \textbf{45.4} & \textbf{40.8} & \textbf{30.6} & \textbf{30.1} & \textbf{31.2} & \textbf{30.7} & \textbf{78.6} \\
\bottomrule
\end{tabular}
\caption{Comparison of existing segmentation methods on the ScanNet200 validation split. Our OneFormer3D sets the new state-of-the art in all segmentation tasks: instance, semantic, and panoptic.}
\label{tab:scannet200}
\end{table*}

On S3DIS dataset, our unified approach demonstrates state-of-the-art results on all segmentation tasks, in both Area-5 and 6-fold cross-validation benchmarks. Here, the most significant gain is achieved in instance segmentation on 6-fold cross-validation, with +1.5 mAP\textsubscript{50} and +1.2 mAP w.r.t. Mask3D. In both benchmarks, we outperform state-of-the-art TD3D and Mask3D in terms of mPrec\textsubscript{50} and mRec\textsubscript{50}. Despite we find these metrics less representative than mAP, we report them to fairly compare with previous methods, and to maintain consistency of the established evaluation protocol.

We also demonstrate the top 3D instance segmentation quality on the ScanNet200 validation split, achieving at least +3 in mAP\textsubscript{25}, mAP\textsubscript{50}, and mAP. To the best of our knowledge, no panoptic segmentation results on ScanNet200 and S3DIS has been reported so far, so we provide our scores as a basis for the future research in this field.

\subsection{Ablation Studies}

\paragraph{Query selection \& disentangled matching.} First, we ablate key novel components of our pipeline on the ScanNet validation split, and report the results in Tab. \ref{tab:matcher}. In this study, we only compare instance segmentation metrics, since both ablated components do not affect semantic segmentation. SPFormer \cite{sun2023spformer} uses random query initialization and Hungarian matching strategy; we evaluate it with the same backbone for a fair comparison. Evidently, our reimplementation with joint instance and semantic training has a minor gain over the baseline. Besides, our query selection scheme does not improve the quality if combined with the baseline bipartite matching scheme. But, the synergy of these two modifications allows for the state-of-the-art results, improving mAP\textsubscript{25}, mAP\textsubscript{50}, and mAP by at least +1.3.

\begin{table}
\centering
\small
\begin{tabular}{llcccc}
\toprule
& QS & Matching & mAP\textsubscript{25} & mAP\textsubscript{50} & mAP \\
\midrule
\multicolumn{3}{l}{\textit{SPFormer, baseline}} & & \\
& & Hungarian & 82.9 & 73.9 & 56.3 \\
\midrule
\multicolumn{3}{l}{\textit{OneFormer3D, ours}} & & \\
& & Hungarian & 84.4 & 75.6 & 58.0 \\
& \cmark & Hungarian & 84.6 & 75.9 & 58.1 \\
& \cmark & Disentangled & 86.4 & 78.1 & 59.3 \\
\bottomrule
\end{tabular}
\caption{Comparison of different query initialization and matching strategies on the ScanNet validation split. QS stands for our query selection.}
\label{tab:matcher}
\end{table}

\paragraph{Pretraining and pooling.} Previous state-of-the-art methods \cite{schult2023mask3d, vu2022softgroup, chen2021hais, sun2023spformer} use pretraining to achieve the highest scores on the S3DIS dataset, as this dataset is fairly small, with 272 scenes in total. Following the best practices, we pre-train our OneFormer3D on ScanNet, which gives significant performance boost: +8.0 mAP\textsubscript{50} and +10.2 mIoU (Tab. \ref{tab:pretraining}). When being pretrained on ScanNet, OneFormer3D and SPFormer demonstrate comparable results. We also leverage a large scale synthetic Structured3D \cite{zheng2020structured3d} dataset for pretraining, which is an order of magnitude larger than ScanNet, with as many as 21835 scenes. In this experiment, benefits from using a larger amount of training data exceed the possible negative effect of a domain gap: the best results are achieved with pretraining on a mixture of real and synthetic data, bringing at least +11.5 in both mAP\textsubscript{50} and mIoU.

Besides, we investigate how our flexible pooling affects the final performance. To this end, we switch it off by replacing superpoints with voxels of 5cm. According to the Tab. \ref{tab:pretraining}, the gain is at least of +1.2 in both mAP\textsubscript{50} and mIoU. Yet, we should mention that superpoint clustering takes almost a half of the entire inference time, so removing it causes at least two times speed-up and eliminates the need to select and tune such an algorithm for each dataset.

\begin{table}
\centering
\small
\begin{tabular}{lccccc}
\toprule
& \multicolumn{2}{c}{Pretrain on} & \multirow{3}{*}{Pooling} & \multirow{3}{*}{\shortstack{Instance\\mAP\textsubscript{50}}} & \multirow{3}{*}{\shortstack{Semantic\\mIoU}} \\
& Scan- & Struc- & & & \\
& Net & tured3D & & & \\
\midrule
\multicolumn{3}{l}{\textit{SPFormer, baseline}} & & & \\
& \cmark & & superpoint & 66.8 & \\
\midrule
\multicolumn{3}{l}{\textit{OneFormer3D, ours}} & & & \\
& & & voxel & 60.5 & 59.1 \\
& \cmark & & voxel & 68.5 & 69.3 \\
& \cmark & & superpoint & 67.1 & 68.1 \\
& & \cmark & voxel & 65.1 & 66.2 \\
& \cmark & \cmark & voxel & \textbf{72.0} & \textbf{72.4} \\
\bottomrule
\end{tabular}
\caption{Ablation study of pretraining and feature pooling on S3DIS Area-5 validation split. We demonstrate the importance of large-scale pretraining on the mixture of real (ScanNet) and synthetic (Structured3D) data.}
\label{tab:pretraining}
\end{table}

\paragraph{Joint training.} Training a single unified model instead of three reduces the training time three times, but, more importantly, it also improves segmentation metrics. As can be seen from Tab. \ref{tab:queries}, instance segmentation accuracy remains unchanged, while the accuracy of semantic predictions grows by as much as +3.4 mIoU. We assume that using a large transformer decoder causes overfitting for a semantic segmentation task, but adding an extra instance segmentation task works as a regularization and reduces the overfitting, hence improving the semantic score. For instance segmentation, the improvement is negligible, mainly because semantic annotations for all classes (except \textit{stuff}) can be derived from instance one, so adds limited new information for model training.

\begin{table}
\centering
\small
\begin{tabular}{ccccc}
\toprule
Instance & Semantic & Instance & Semantic & Panoptic \\
queries & queries & mAP\textsubscript{50} & mIoU & PQ \\
\midrule
\cmark & & 78.1 & & \\
& \cmark & & 72.8 & \\
\cmark & \cmark & \textbf{78.1} & \textbf{76.6} & \textbf{71.2} \\
\bottomrule
\end{tabular}
\caption{Importance of joint training with instance and semantic queries on ScanNet validation split. OneFormer3D not only allows panoptic segmentation for free, but also improves semantic segmentation metrics.}
\label{tab:queries}
\end{table}

\section{Conclusion}

In this paper, we proposed a novel transformer-based framework, OneFormer3D, that unifies three 3D point cloud segmentations tasks: instance, semantic, and panoptic. Trained only once on a panoptic dataset, OneFormer3D consistently outperforms existing segmentation approaches -- even though they are trained separately on each task. We also identified the weaknesses of existing transformer-based 3D instance segmentation methods, and addressed them with a novel query selection and disentangled matching strategies. In extensive experiments on ScanNet, ScanNet200, and S3DIS, OneFormer3D established a new state-of-the-art in all three 3D segmentation tasks.
\appendix

\section{Per-category Scores}

Since segmentation tasks are severely imbalanced in terms of categories, an averaged score might shadow some crucial performance issues. To provide a complete picture, 
we report 3D panoptic segmentation scores on the ScanNet validation split and on the S3DIS Area-5 split in Tab. \ref{tab:per_class_scannet_panoptic} and \ref{tab:per_class_s3dis_panoptic}, respectively. Besides, per-category 3D instance segmentation scores on the ScanNet test split are listed in Tab. \ref{tab:per_class_scannet_instance}. Evidently, OneFormer3D segments every single category more precisely than competitors on the ScanNet validation split. Panoptic segmentation scores on S3DIS have never been reported so far, so we establish a baseline for future research. On the ScanNet test split, our method outperforms others in segmenting objects of 11 out of 18 categories.

\begin{table*}[h!]
\centering
\resizebox{\textwidth}{!}{
\begin{tabular}{lccccccccccccccccccccc}
\toprule
\multirow{3}{*}{Method} & \multicolumn{1}{l}{\multirow{3}{*}{PQ}} & \multirow{3}{*}{\rotatebox{90}{wall}} & \multirow{3}{*}{\rotatebox{90}{floor}} & \multirow{3}{*}{\rotatebox{90}{cabinet}} & \multirow{3}{*}{\rotatebox{90}{bed}} & \multirow{3}{*}{\rotatebox{90}{chair}} & \multirow{3}{*}{\rotatebox{90}{sofa}} & \multirow{3}{*}{\rotatebox{90}{table}} & \multirow{3}{*}{\rotatebox{90}{door}} & \multirow{3}{*}{\rotatebox{90}{window}} & \multirow{3}{*}{\rotatebox{90}{bkshf}} & \multirow{3}{*}{\rotatebox{90}{picture}} & \multirow{3}{*}{\rotatebox{90}{counter}} & \multirow{3}{*}{\rotatebox{90}{desk}} & \multirow{3}{*}{\rotatebox{90}{curtain}} & \multirow{3}{*}{\rotatebox{90}{fridge}} & \multirow{3}{*}{\rotatebox{90}{s. cur.}} & \multirow{3}{*}{\rotatebox{90}{toilet}} & \multirow{3}{*}{\rotatebox{90}{sink}} & \multirow{3}{*}{\rotatebox{90}{bath}} & \multirow{3}{*}{\rotatebox{90}{other}} \\ 
& \multicolumn{1}{l}{} & \multicolumn{1}{l}{} & \multicolumn{1}{l}{} & \multicolumn{1}{l}{} & \multicolumn{1}{l}{} & \multicolumn{1}{l}{} & \multicolumn{1}{l}{} & \multicolumn{1}{l}{} & \multicolumn{1}{l}{} & \multicolumn{1}{l}{} & \multicolumn{1}{l}{} & \multicolumn{1}{l}{} & \multicolumn{1}{l}{} & \multicolumn{1}{l}{} & \multicolumn{1}{l}{} & \multicolumn{1}{l}{} & \multicolumn{1}{l}{} & \multicolumn{1}{l}{} & \multicolumn{1}{l}{} \\
& \multicolumn{1}{l}{} & \multicolumn{1}{l}{} & \multicolumn{1}{l}{} & \multicolumn{1}{l}{} & \multicolumn{1}{l}{} & \multicolumn{1}{l}{} & \multicolumn{1}{l}{} & \multicolumn{1}{l}{} & \multicolumn{1}{l}{} & \multicolumn{1}{l}{} & \multicolumn{1}{l}{} & \multicolumn{1}{l}{} & \multicolumn{1}{l}{} & \multicolumn{1}{l}{} & \multicolumn{1}{l}{} & \multicolumn{1}{l}{} & \multicolumn{1}{l}{} & \multicolumn{1}{l}{} & \multicolumn{1}{l}{} \\
\midrule
SceneGraphFusion~\cite{wu2021scenegraphfusion} & 31.5 & 67.6 & 25.4 & 13.9 & 22.2 & 47.2 & 10.5 & 16.4 & 12.6 & 26.4 & 56.4 & 22.9 & 31.3 & 28.0 & 38.3 & 38.0 & 32.3 & 34.8 & 63.2 & 30.4 & 11.7 \\
PanopticFusion~\cite{narita2019panopticfusion} & 33.5 & 40.4 & 76.4 & 23.8 & 35.8 & 46.7 & 42.1 & 34.8 & 18.0 & 19.3 & 16.4 & 26.4 & 10.4 & 16.1 & 16.6 & 39.5 & 36.3 & 76.1 & 36.7 & 31.0 & 27.7 \\
TUPPer-Map~\cite{yang2021tuppermap} & 50.2 & 68.5 & 74.6 & 47.1 & 60.3 & 45.8 & 49.6 & 52.5 & 38.1 & 38.7 & 53.5 & 42.0 & 38.8 & 44.6 & 32.6 & 47.5 & 52.3 & 74.5 & 45.5 & 57.4 & 39.9 \\
\textbf{OneFormer3D} & \textbf{71.2} & \textbf{78.9} & \textbf{94.9} & \textbf{60.9} & \textbf{80.4} & \textbf{88.8} & \textbf{74.4} & \textbf{74.4} & \textbf{61.5} & \textbf{58.9} & \textbf{55.2} & \textbf{57.1} & \textbf{55.8} & \textbf{65.7} & \textbf{62.5} & \textbf{63.3} & \textbf{71.7} & \textbf{95.9} & \textbf{73.7} & \textbf{85.5} & \textbf{65.2} \\
\bottomrule
\end{tabular}}
\caption{Per-class 3D panoptic segmentation PQ scores on the ScanNet validation split.}
\label{tab:per_class_scannet_panoptic}
\end{table*}

\begin{table*}[h!]
\centering
\small
\begin{tabular}{lcccccccccccccc}
\toprule
\multirow{3}{*}{Method} & \multicolumn{1}{l}{\multirow{3}{*}{PQ}} & \multirow{3}{*}{\rotatebox{90}{ceiling}} & \multirow{3}{*}{\rotatebox{90}{floor}} & \multirow{3}{*}{\rotatebox{90}{wall}} & \multirow{3}{*}{\rotatebox{90}{beam}} & \multirow{3}{*}{\rotatebox{90}{column}} & \multirow{3}{*}{\rotatebox{90}{window}} & \multirow{3}{*}{\rotatebox{90}{door}} & \multirow{3}{*}{\rotatebox{90}{table}} & \multirow{3}{*}{\rotatebox{90}{chair}}  & \multirow{3}{*}{\rotatebox{90}{sofa}} & \multirow{3}{*}{\rotatebox{90}{b. case}} & \multirow{3}{*}{\rotatebox{90}{board}} & \multirow{3}{*}{\rotatebox{90}{clutter}}\\ 
& \multicolumn{1}{l}{} & \multicolumn{1}{l}{} & \multicolumn{1}{l}{} & \multicolumn{1}{l}{} & \multicolumn{1}{l}{} & \multicolumn{1}{l}{} & \multicolumn{1}{l}{} & \multicolumn{1}{l}{} & \multicolumn{1}{l}{} & \multicolumn{1}{l}{} & \multicolumn{1}{l}{} & \multicolumn{1}{l}{} \\
& \multicolumn{1}{l}{} & \multicolumn{1}{l}{} & \multicolumn{1}{l}{} & \multicolumn{1}{l}{} & \multicolumn{1}{l}{} & \multicolumn{1}{l}{} & \multicolumn{1}{l}{} & \multicolumn{1}{l}{} & \multicolumn{1}{l}{} & \multicolumn{1}{l}{} & \multicolumn{1}{l}{} & \multicolumn{1}{l}{} \\
\midrule
OneFormer3D & 62.2 & 92.0 & 96.5 & 81.5 & 0.0 & 40.9 & 66.2 & 81.4 & 43.9 & 87.0 & 48.5 & 46.0 & 81.3 & 43.9 \\
\bottomrule
\end{tabular}
\caption{Per-class 3D panoptic segmentation PQ scores on the S3DIS Area-5 split.}
\label{tab:per_class_s3dis_panoptic}
\end{table*}

\begin{table*}[h!]
\centering
\resizebox{\textwidth}{!}{
\begin{tabular}{lccccccccccccccccccc}
\toprule
\multirow{3}{*}{Method} & \multicolumn{1}{l}{\multirow{3}{*}{mAP\textsubscript{50}}} & \multirow{3}{*}{\rotatebox{90}{bath}} & \multirow{3}{*}{\rotatebox{90}{bed}} & \multirow{3}{*}{\rotatebox{90}{bkshf}} & \multirow{3}{*}{\rotatebox{90}{cabinet}} & \multirow{3}{*}{\rotatebox{90}{chair}} & \multirow{3}{*}{\rotatebox{90}{counter}} & \multirow{3}{*}{\rotatebox{90}{curtain}} & \multirow{3}{*}{\rotatebox{90}{desk}} & \multirow{3}{*}{\rotatebox{90}{door}} & \multirow{3}{*}{\rotatebox{90}{other}} & \multirow{3}{*}{\rotatebox{90}{picture}} & \multirow{3}{*}{\rotatebox{90}{fridge}} & \multirow{3}{*}{\rotatebox{90}{s. cur.}} & \multirow{3}{*}{\rotatebox{90}{sink}} & \multirow{3}{*}{\rotatebox{90}{sofa}} & \multirow{3}{*}{\rotatebox{90}{table}} & \multirow{3}{*}{\rotatebox{90}{toilet}} & \multirow{3}{*}{\rotatebox{90}{window}} \\ 
& \multicolumn{1}{l}{} & \multicolumn{1}{l}{} & \multicolumn{1}{l}{} & \multicolumn{1}{l}{} & \multicolumn{1}{l}{} & \multicolumn{1}{l}{} & \multicolumn{1}{l}{} & \multicolumn{1}{l}{} & \multicolumn{1}{l}{} & \multicolumn{1}{l}{} & \multicolumn{1}{l}{} & \multicolumn{1}{l}{} & \multicolumn{1}{l}{} & \multicolumn{1}{l}{} & \multicolumn{1}{l}{} & \multicolumn{1}{l}{} & \multicolumn{1}{l}{} & \multicolumn{1}{l}{} & \multicolumn{1}{l}{} \\
& \multicolumn{1}{l}{} & \multicolumn{1}{l}{} & \multicolumn{1}{l}{} & \multicolumn{1}{l}{} & \multicolumn{1}{l}{} & \multicolumn{1}{l}{} & \multicolumn{1}{l}{} & \multicolumn{1}{l}{} & \multicolumn{1}{l}{} & \multicolumn{1}{l}{} & \multicolumn{1}{l}{} & \multicolumn{1}{l}{} & \multicolumn{1}{l}{} & \multicolumn{1}{l}{} & \multicolumn{1}{l}{} & \multicolumn{1}{l}{} & \multicolumn{1}{l}{} & \multicolumn{1}{l}{} & \multicolumn{1}{l}{} \\
\midrule
NeuralBF~\cite{sun2023neuralbf} & 55.5 & 66.7 & 89.6 & 84.3 & 51.7 & 75.1 & 2.9 & 51.9 & 41.4 & 43.9 & 46.5 & 0.0 & 48.4 & 85.7 & 28.7 & 69.3 & 65.1 & 100 & 48.5 \\
PointGroup~\cite{jiang2020pointgroup} & 63.6 & 100 & 76.5 & 62.4 & 50.5 & 79.7 & 11.6 & 69.6 & 38.4 & 44.1 & 55.9 & 47.6 & 59.6 & 100 & 66.6 & 75.6 & 55.6 & 99.7 & 51.3 \\
DyCo3D~\cite{he2021dyco3d} & 64.1 & 100 & 84.1 & 89.3 & 53.1 & 80.2 & 11.5 & 58.8 & 44.8 & 43.8 & 53.7 & 43.0 & 55.0 & 85.7 & 53.4 & 76.4 & 65.7 & 98.7 & 56.8 \\
SSTNet~\cite{liang2021sstnet} & 69.8 & 100 & 69.7 & 88.8 & 55.6 & 80.3 & 38.7 & 62.6 & 41.7 & 55.6 & 58.5 & 70.2 & 60.0 & 100 & 82.4 & 72.0 & 69.2 & 100 & 50.9 \\
HAIS~\cite{chen2021hais} & 69.9 & 100 & 84.9 & 82.0 & 67.5 & 80.8 & 27.9 & 75.7 & 46.5 & 51.7 & 59.6 & 55.9 & 60.0 & 100 & 65.4 & 76.7 & 67.6 & 99.4 & 56.0 \\
DKNet~\cite{wu2022dknet} & 71.8 & 100 & 81.4 & 78.2 & 61.9 & 87.2 & 22.4 & 75.1 & 56.9 & 67.7 & 58.5 & 72.4 & 63.3 & 98.1 & 51.5 & 81.9 & 73.6 & 100 & 61.7 \\
TD3D~\cite{kolodiazhnyi2023td3d} & 75.1 & 100 & 77.4 & 86.7 & 62.1 & 93.4 & 40.4 & 70.6 & 81.2 & 60.5 & 63.3 & 62.6 & 69.0 & 100 & 64.0 & 82.0 & 77.7 & 100 & 61.2 \\
ISBNet~\cite{ngo2023isbnet} & 75.7 & 100 & 90.4 & 73.1 & 67.8 & 89.5 & 45.8 & 64.4 & 67.0 & 71.0 & 62.0 & 73.2 & 65.0 & 100 & 75.6 & 77.8 & 77.9 & 100 & 61.4 \\
SPFormer~\cite{sun2023spformer} & 77.0 & 90.3 & 90.3 & 80.6 & 60.9 & 88.6 & 56.8 & 81.5 & 70.5 & 71.1 & 65.5 & 65.2 & 68.5 & 100 & 78.9 & 80.9 & 77.6 & 100 & 58.3 \\
Mask3D~\cite{schult2023mask3d} & 78.0 & 100 & 78.6 & 71.6 & 69.6 & 88.5 & 50.0 & 71.4 & 81.0 & 67.2 & 71.5 & 67.9 & 80.9 & 100 & 83.1 & 83.3 & 78.7 & 100 & 60.2 \\
\textbf{OneFormer3D} & \textbf{80.1} & 100 & 97.3 & 90.9 & 69.8 & 92.8 & 58.2 & 66.8 & 68.5 & 78.0 & 68.7 & 69.8 & 70.2 & 100 & 79.4 & 90.0 & 78.4 & 98.6 & 63.5 \\
\bottomrule
\end{tabular}}
\caption{Per-class 3D instance segmentation mAP\textsubscript{50} scores on the ScanNet hidden test split at 17 Nov. 2023.}
\label{tab:per_class_scannet_instance}
\end{table*}

\section{Performance}


To provide a comprehensive overview of the proposed method, we also conduct a detailed performance analysis. Specifically, we decompose our method into several self-sufficient and replaceable components: creating superpoints, extracting 3D features with a sparse 3D CNN, flexible pooling, and running a query decoder. We run a profiler to measure the time required for each component to proceed. Similarly, we identify components of competing approaches, and report the inference time component-wise in Tab. \ref{tab:profiler}. The runtime is measured on the same RTX 3090 GPU. Compared with the SPFormer baseline, OneFormer3D processes a few additional queries for semantic segmentation, and uses another initialization strategy for instance queries. The computation overhead is though minor, causing a less than 3\% increase of inference time. Overall, we can claim, that OneFormer3D is on par of SPFormer, which is the fastest among the profiled approaches. 

\begin{table*}[h!]
\centering
\small
\begin{tabular}{llcccc}
\toprule
\multirow{2}{*}{Method} & \multirow{2}{*}{Component} & \multirow{2}{*}{Device} & \multirow{2}{*}{\shortstack{Component\\time, ms}} & \multirow{2}{*}{\shortstack{Total\\time, ms}} & \multirow{2}{*}{mAP\textsubscript{50}} \\
& & & & & \\
\midrule
\multirow{3}{*}{PointGroup~\cite{jiang2020pointgroup}} & Backbone & GPU & 48 & \multirow{3}{*}{372} & \multirow{3}{*}{56.7} \\
& Grouping & GPU+CPU & 218 & & \\
& ScoreNet & GPU & 106 & & \\
\midrule
\multirow{4}{*}{SSTNet~\cite{liang2021sstnet}} & Superpoint extraction & CPU & 168 & \multirow{4}{*}{400} & \multirow{4}{*}{64.3} \\
& Backbone & GPU & 26 & & \\
& Tree Network & GPU+CPU & 148 & & \\
& ScoreNet & GPU & 58 & & \\
\midrule
\multirow{3}{*}{HAIS~\cite{chen2021hais}} & Backbone & GPU & 50 & \multirow{3}{*}{256} & \multirow{3}{*}{64.4} \\
& Hierarchical aggregation & GPU+CPU & 116 & & \\
& Intra-instance refinement & GPU & 90 & & \\
\midrule
\multirow{3}{*}{SoftGroup~\cite{vu2022softgroup}} & Backbone & GPU & 48 & \multirow{3}{*}{266} & \multirow{3}{*}{67.6} \\
& Soft grouping & GPU+CPU & 121 & & \\
& Top-down refinement & GPU & 97 & & \\
\midrule
\multirow{3}{*}{\shortstack[l]{Mask3D~\cite{schult2023mask3d}\\w/o clustering}} & Backbone & GPU & 106 & \multirow{3}{*}{221} & \multirow{3}{*}{73.0} \\
& Mask module & GPU & 100 & & \\
& Query refinement & GPU & 15 & & \\
\midrule
\multirow{4}{*}{Mask3D~\cite{schult2023mask3d}} & Backbone & GPU & 106 & \multirow{4}{*}{19851} & \multirow{4}{*}{73.7} \\
& Mask module & GPU & 100 & & \\
& Query refinement & GPU & 15 & & \\
& DBSCAN clustering & CPU & 19630 & & \\  
\midrule
\multirow{4}{*}{SPFormer~\cite{sun2023spformer}} & Superpoint extraction & CPU & 168 & \multirow{4}{*}{215} & \multirow{4}{*}{73.9} \\
& Backbone & GPU & 26 & & \\
& Superpoint pooling & GPU & 4 & & \\
& Query decoder & GPU & 17 & & \\  
\midrule
\multirow{4}{*}{\textbf{OneFormer3D}} & Superpoint extraction & CPU & 168 & \multirow{4}{*}{221} & \multirow{4}{*}{78.1} \\
& Backbone & GPU & 26 & & \\
& Superpoint pooling & GPU & 4 & & \\
& Query decoder & GPU & 23 & & \\  
\bottomrule
\end{tabular} 
\caption{The inference time and instance segmentation accuracy on the ScanNet validation split. We show comparable inference time to the fastest SPFormer ~\cite{sun2023spformer}, being significantly more accurate than all existing methods.}
\label{tab:profiler}
\end{table*}

\section{Qualitative Results}

\begin{figure*}
\centering
\setlength{\tabcolsep}{0pt}
\begin{tabular}{ccccc}
Input & Ground Truth & Instance & Semantic & Panoptic \\
\includegraphics[width=0.19\linewidth]{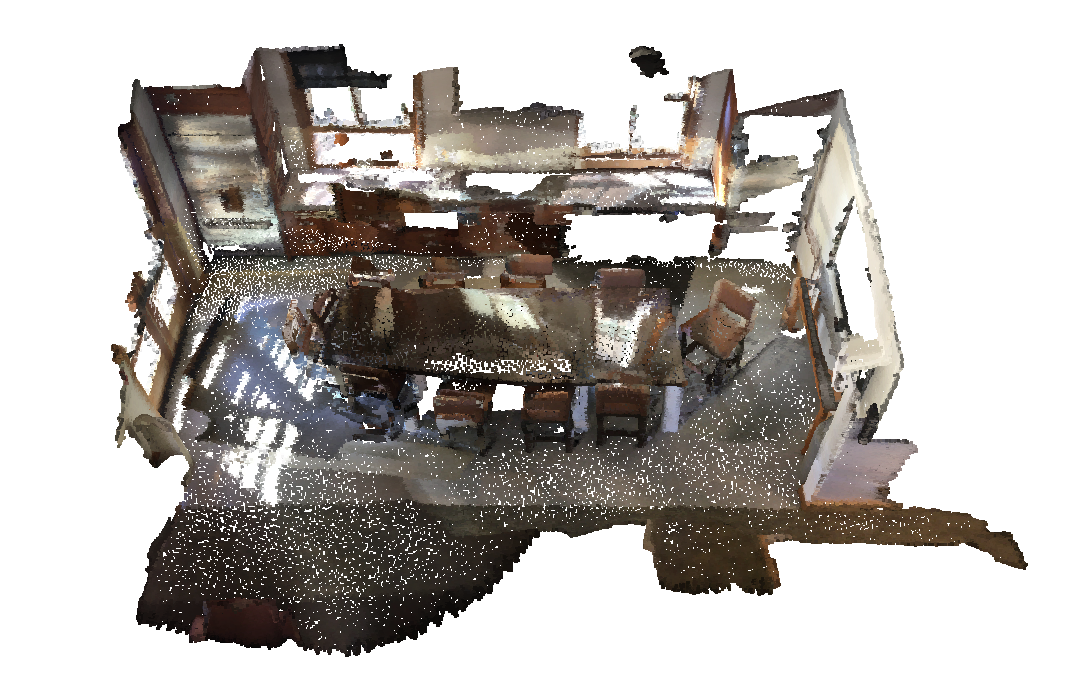} &
\includegraphics[width=0.19\linewidth]{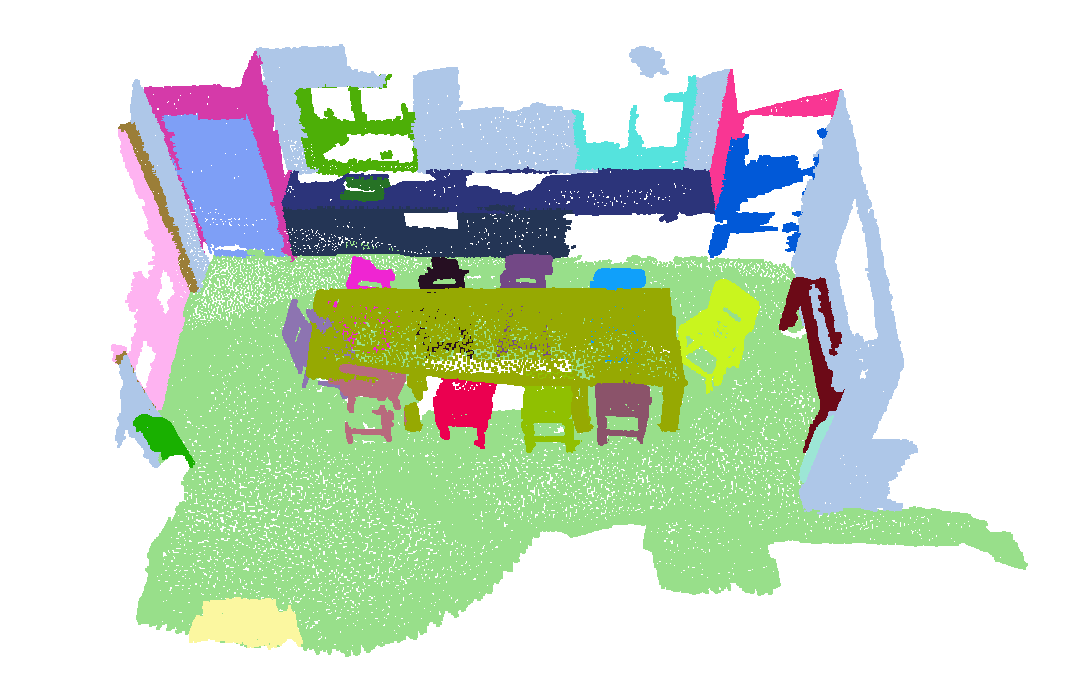} & 
\includegraphics[width=0.19\linewidth]{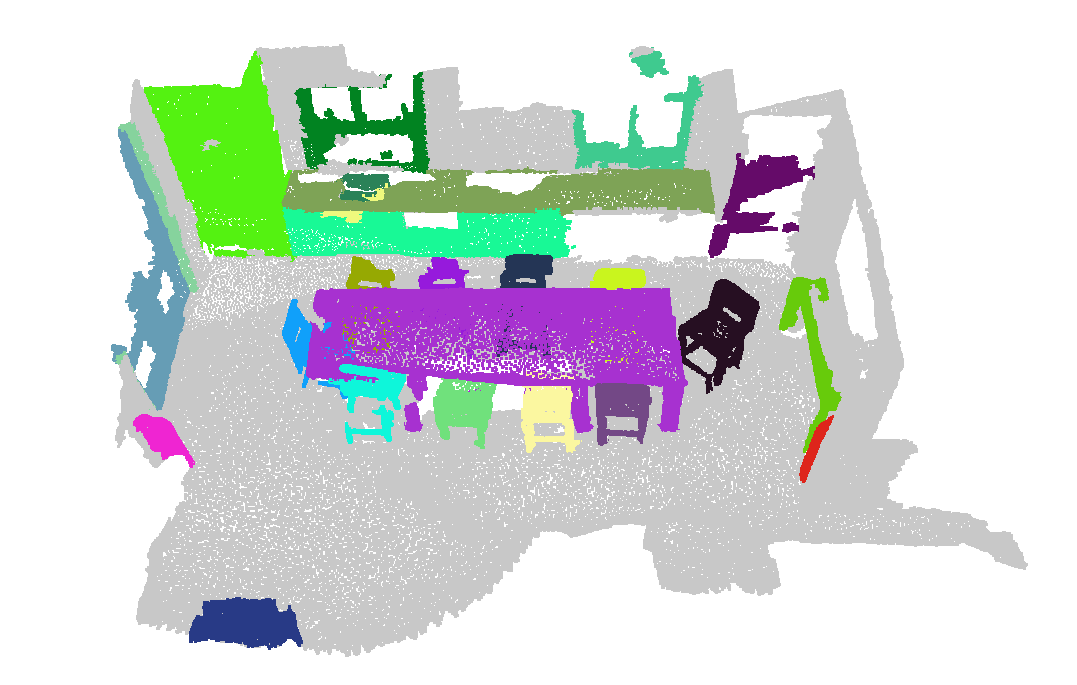} & 
\includegraphics[width=0.19\linewidth]{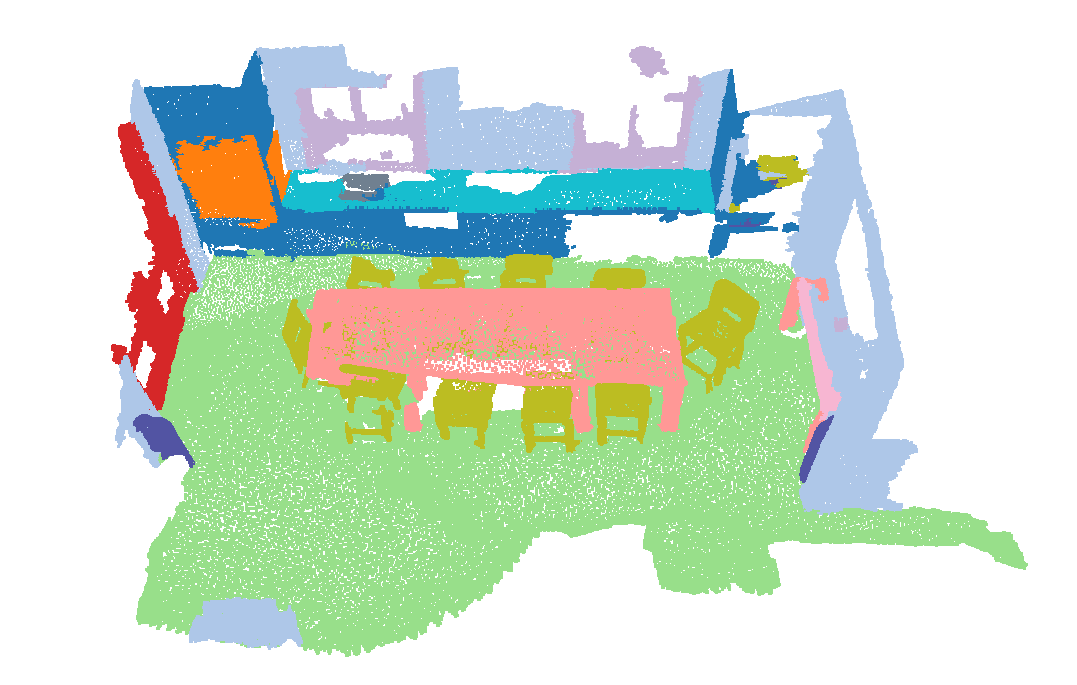} & 
\includegraphics[width=0.19\linewidth]{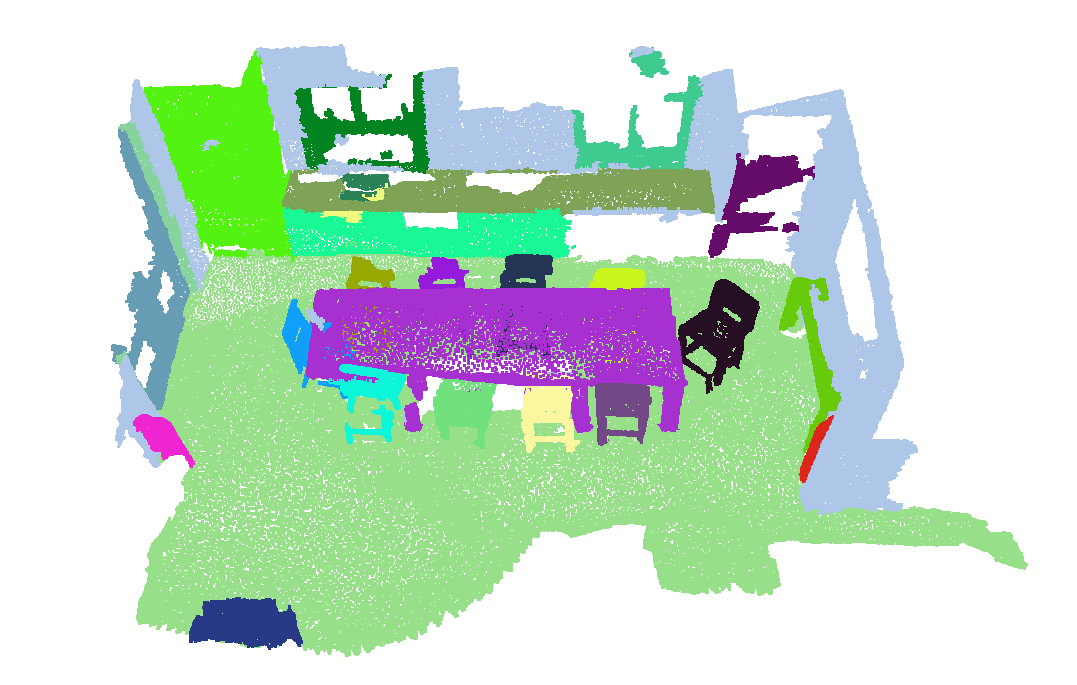} \\
\includegraphics[width=0.19\linewidth]{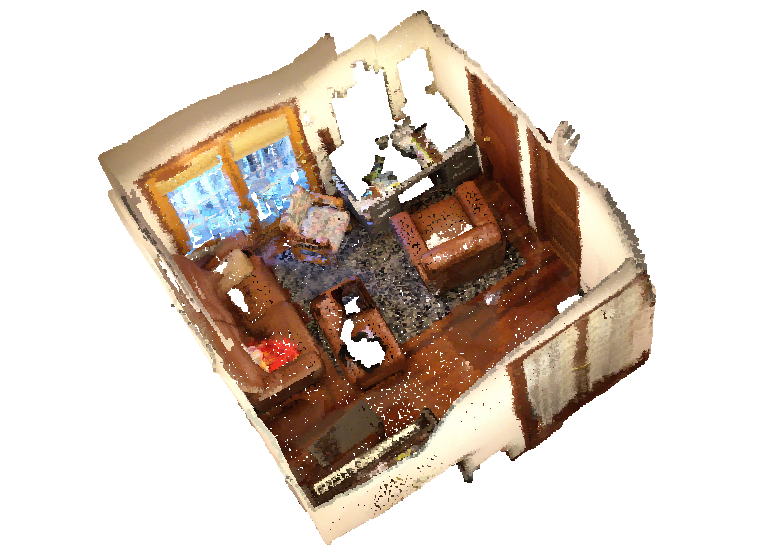} &
\includegraphics[width=0.19\linewidth]{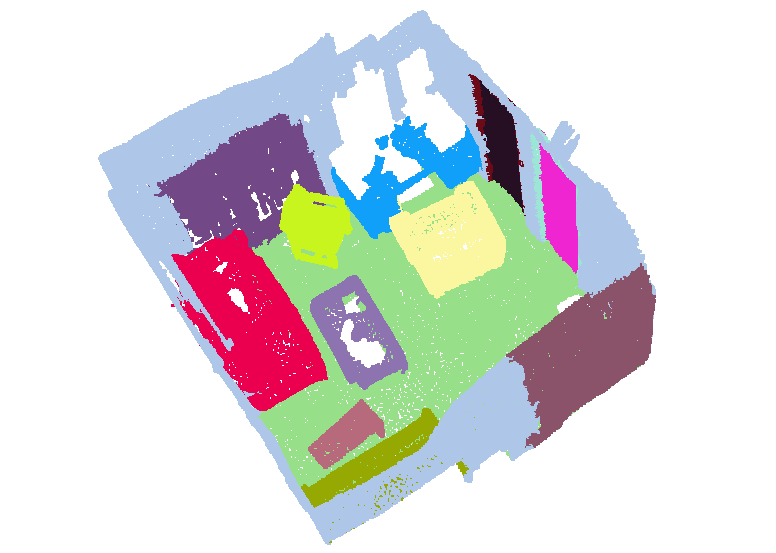} & 
\includegraphics[width=0.19\linewidth]{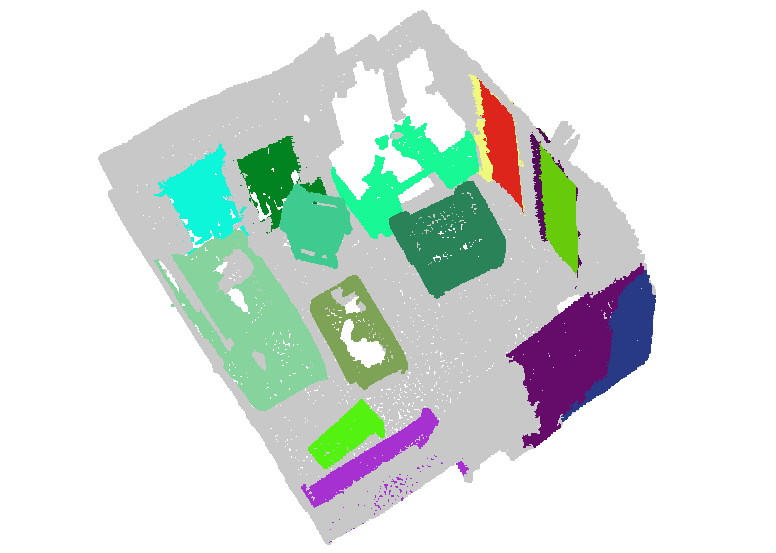} & 
\includegraphics[width=0.19\linewidth]{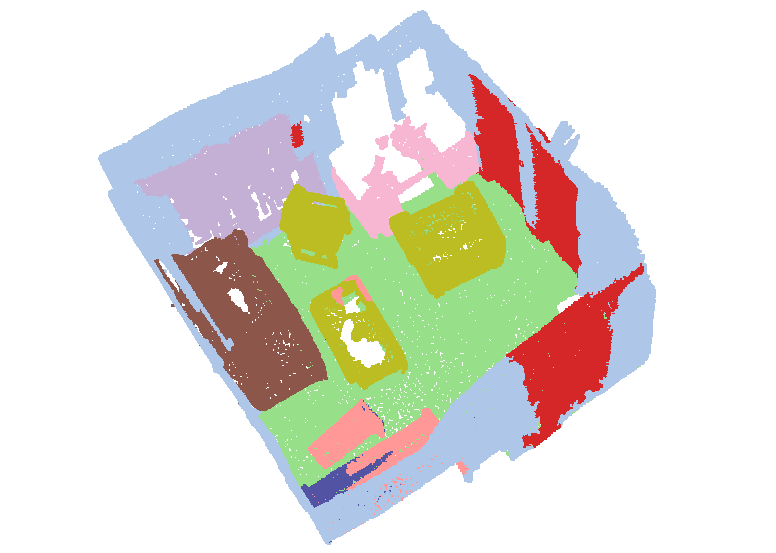} & 
\includegraphics[width=0.19\linewidth]{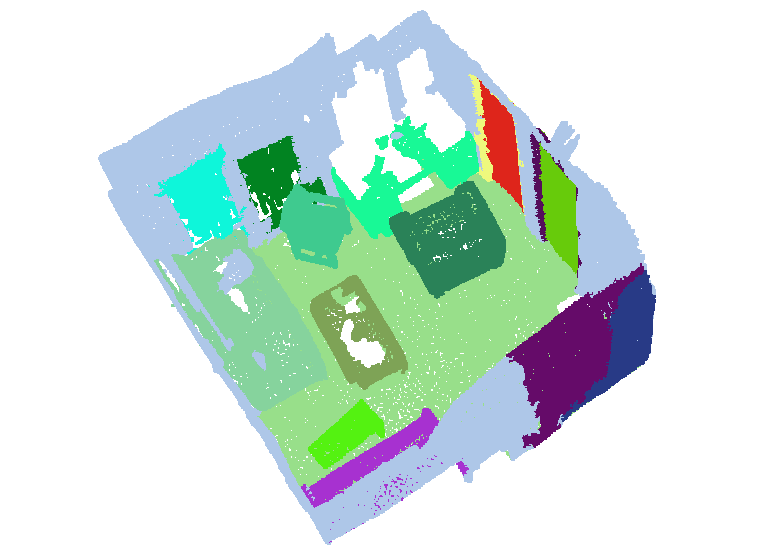} \\
\includegraphics[width=0.19\linewidth]{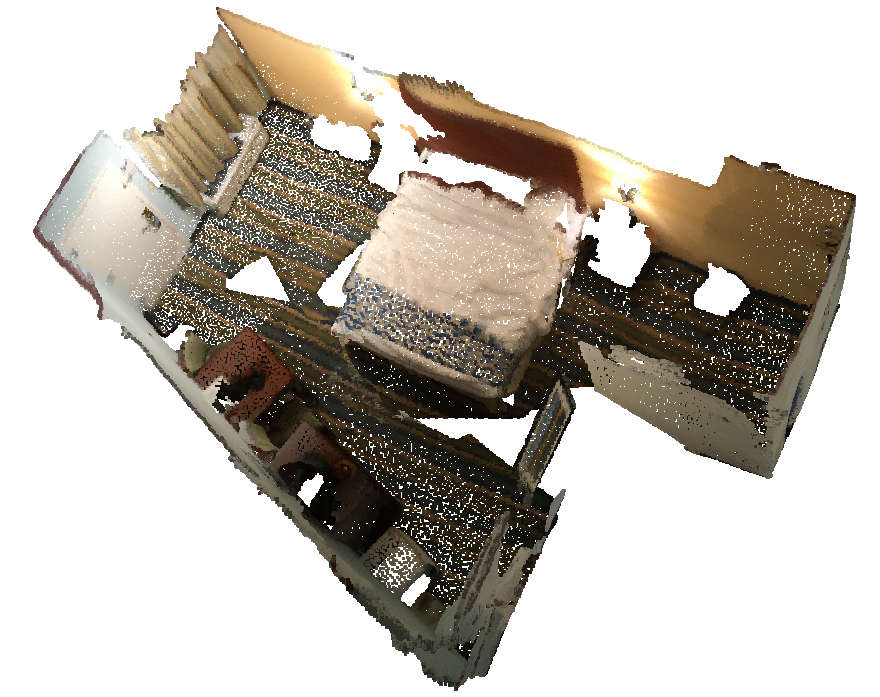} &
\includegraphics[width=0.19\linewidth]{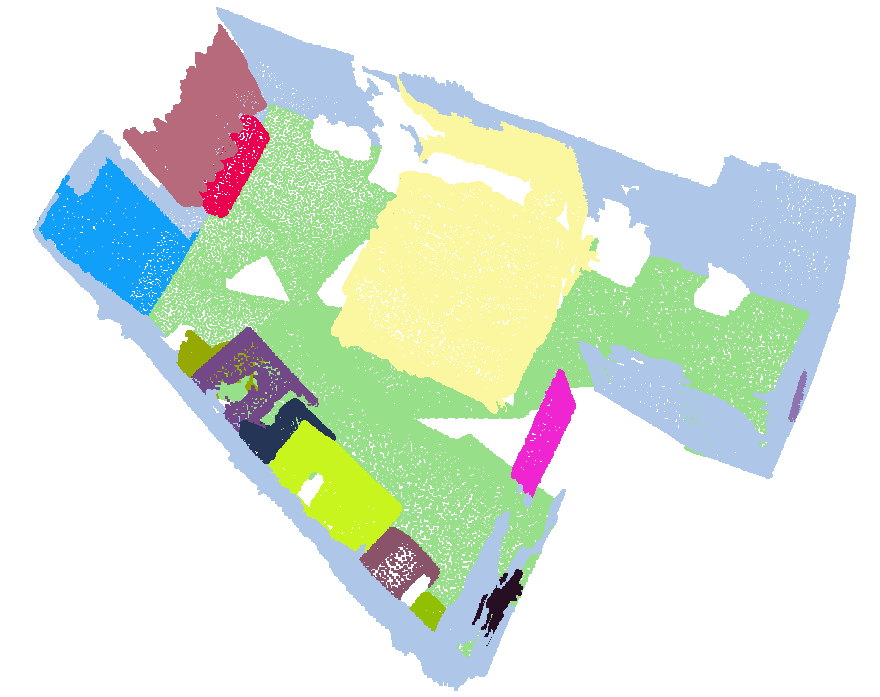} & 
\includegraphics[width=0.19\linewidth]{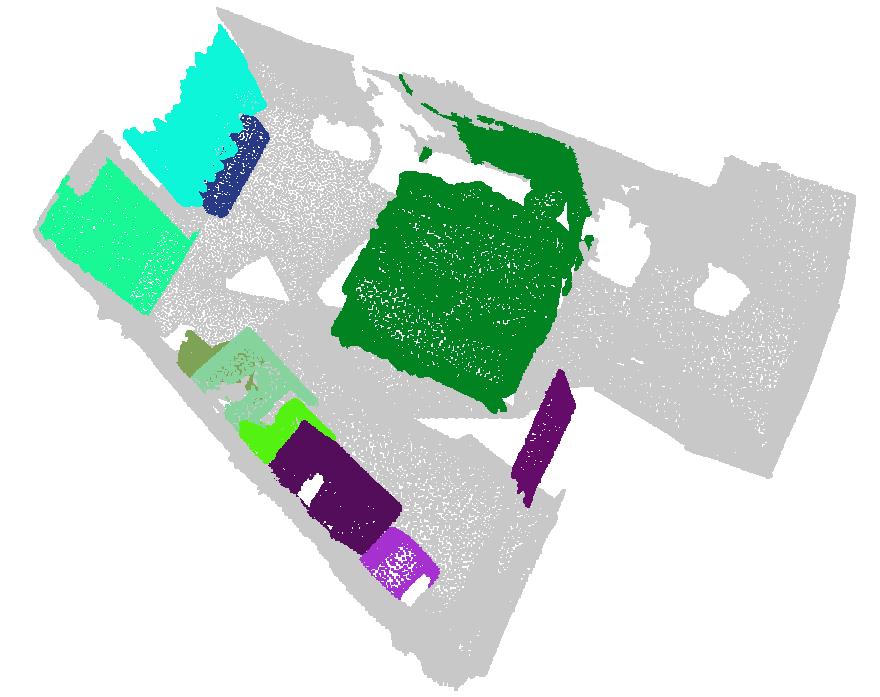} & 
\includegraphics[width=0.19\linewidth]{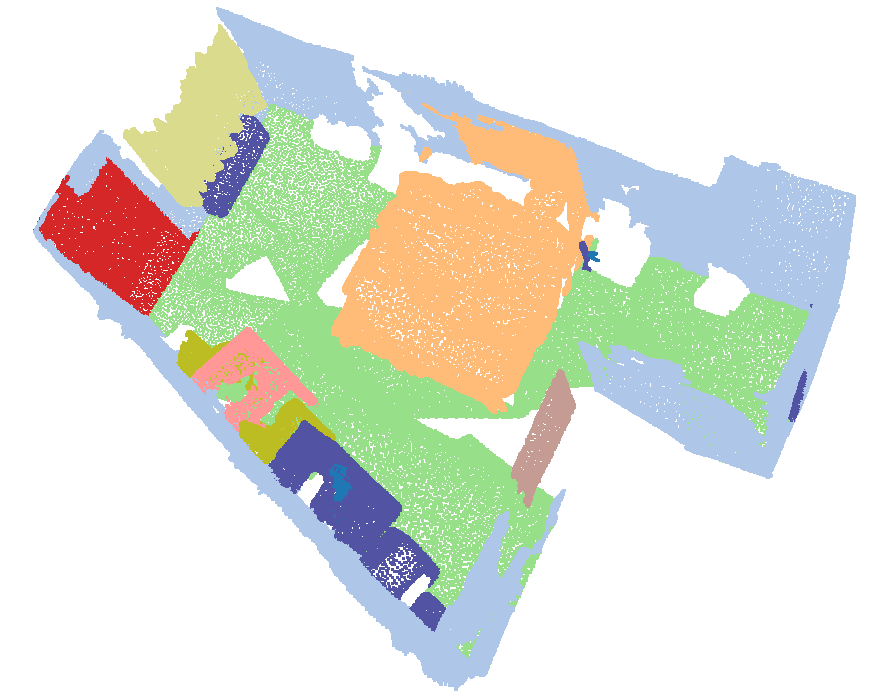} & 
\includegraphics[width=0.19\linewidth]{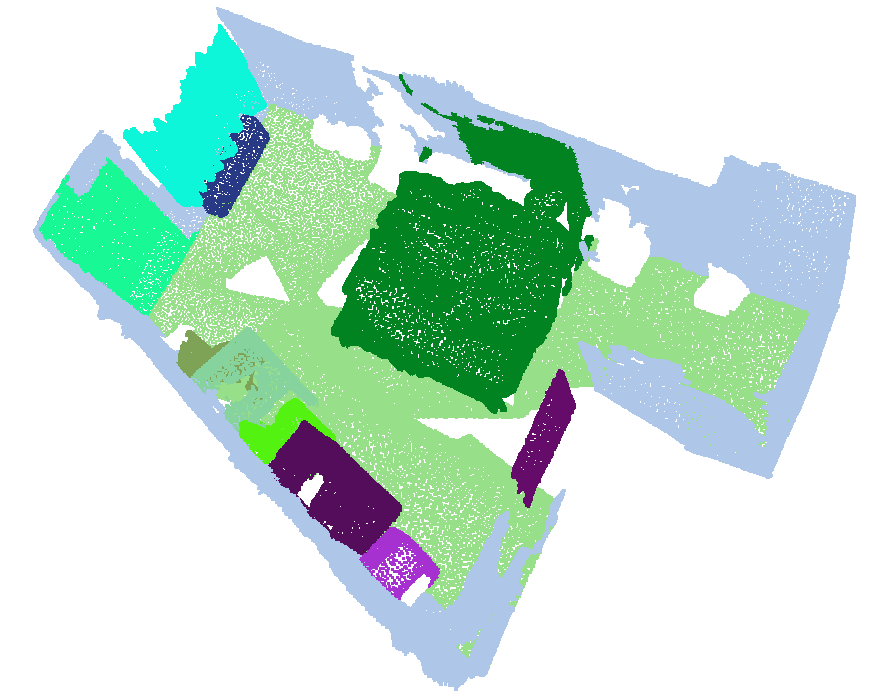} \\
\includegraphics[width=0.19\linewidth]{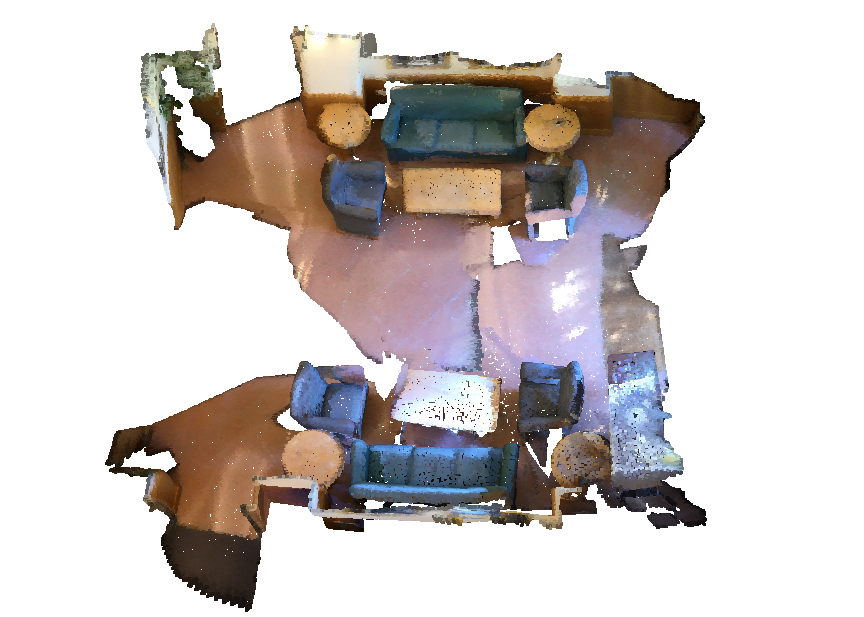} &
\includegraphics[width=0.19\linewidth]{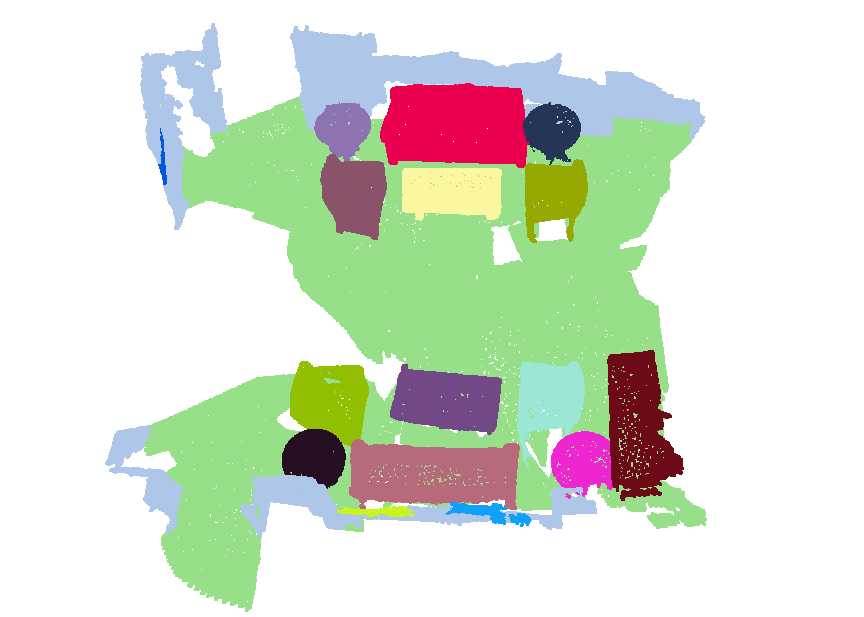} & 
\includegraphics[width=0.19\linewidth]{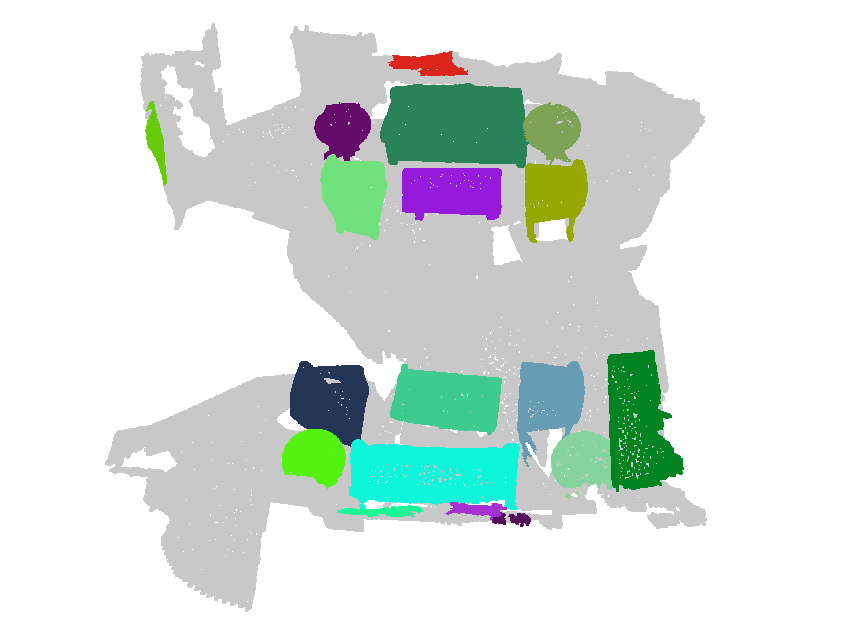} & 
\includegraphics[width=0.19\linewidth]{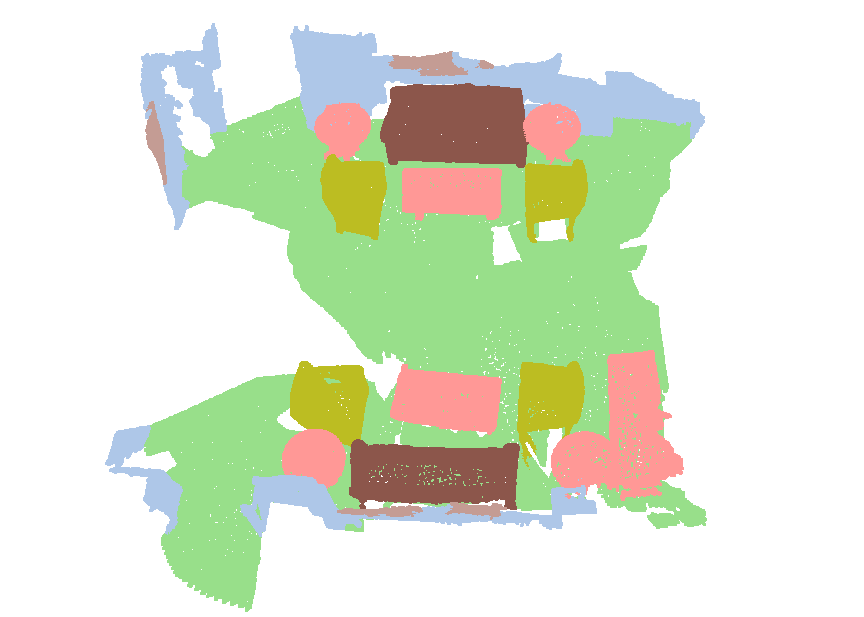} & 
\includegraphics[width=0.19\linewidth]{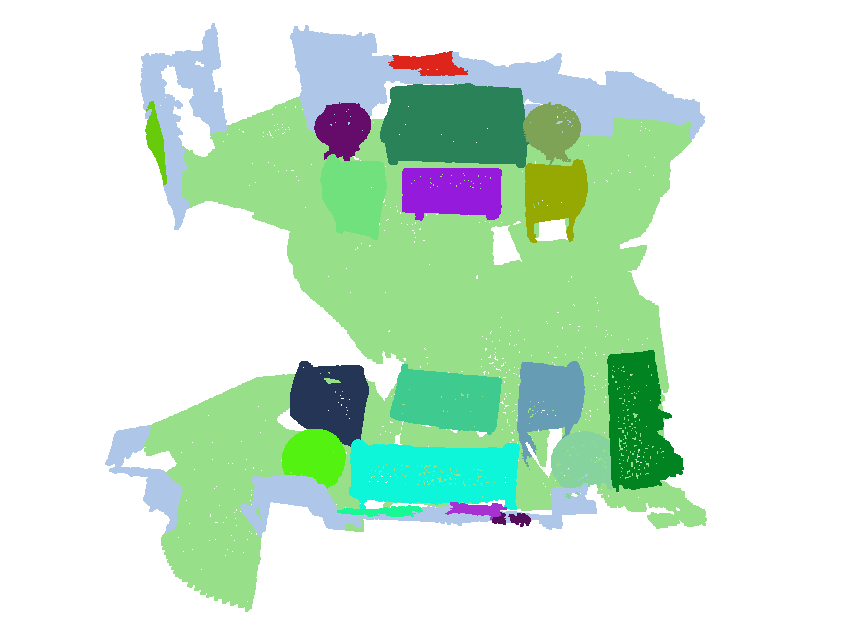} \\
\end{tabular}
\caption{OneFormer3D predictions on ScanNet validation split. Left to right: an input point cloud, a ground truth panoptic mask, predicted 3D instance, 3D semantic, and 3D panoptic segmentation masks.}
\label{fig:visualition_scannet}
\end{figure*}

\begin{figure*}
\centering
\setlength{\tabcolsep}{1pt}
\renewcommand{\arraystretch}{1.5}
\begin{tabular}{ccccc}
Input & Ground Truth & Instance & Semantic & Panoptic \\
\includegraphics[width=0.19\linewidth]{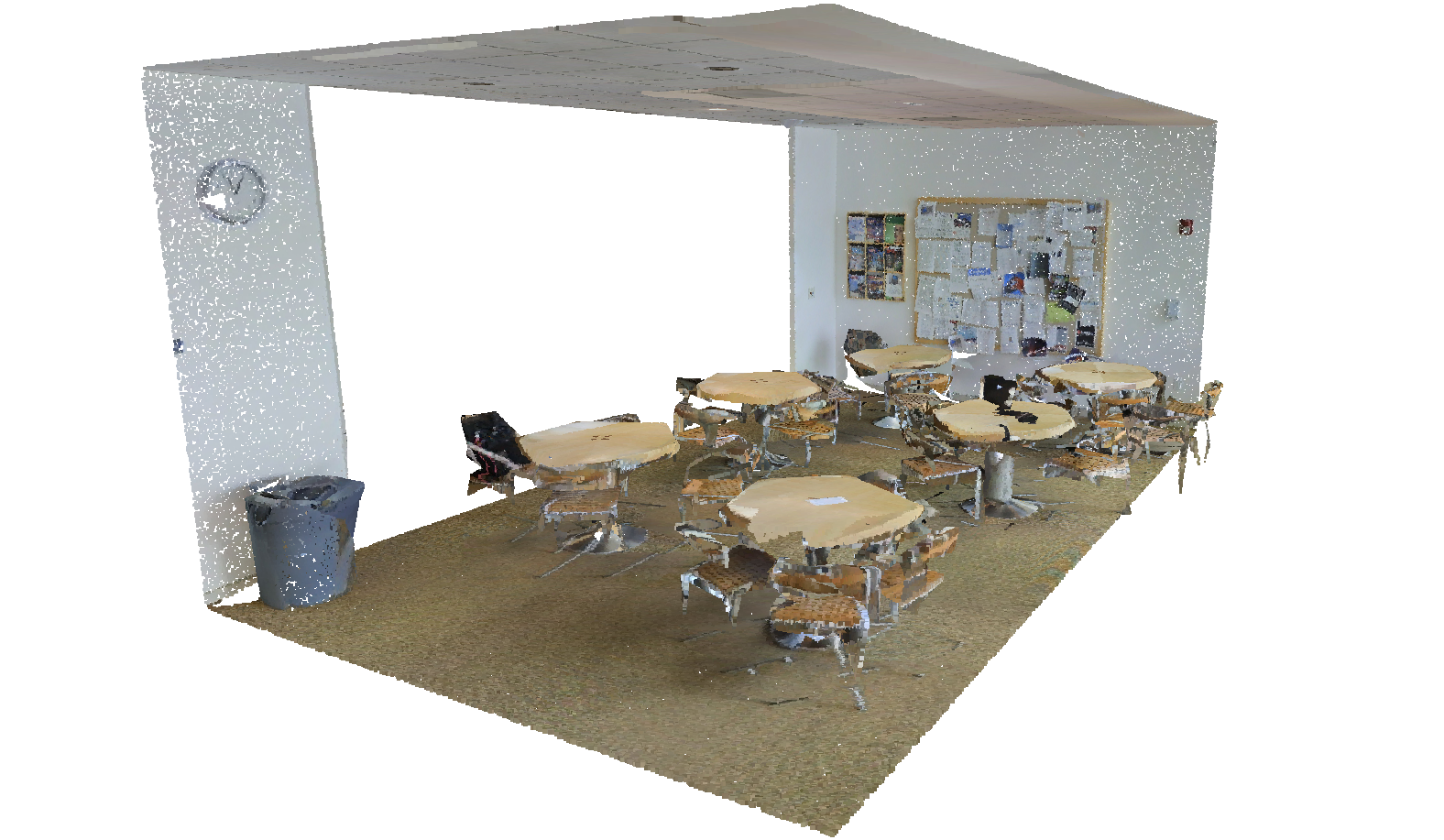} &
\includegraphics[width=0.19\linewidth]{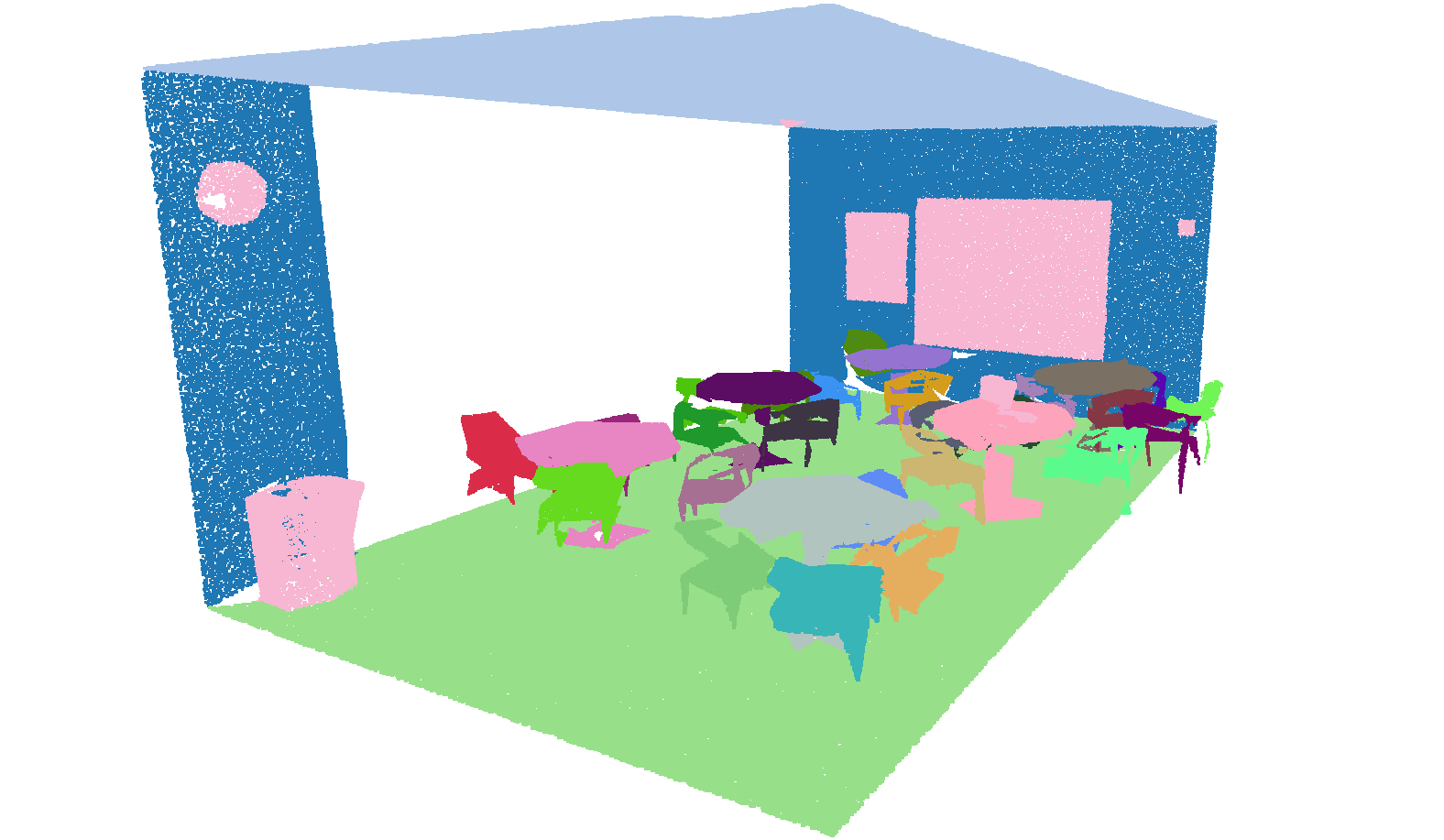} & 
\includegraphics[width=0.19\linewidth]{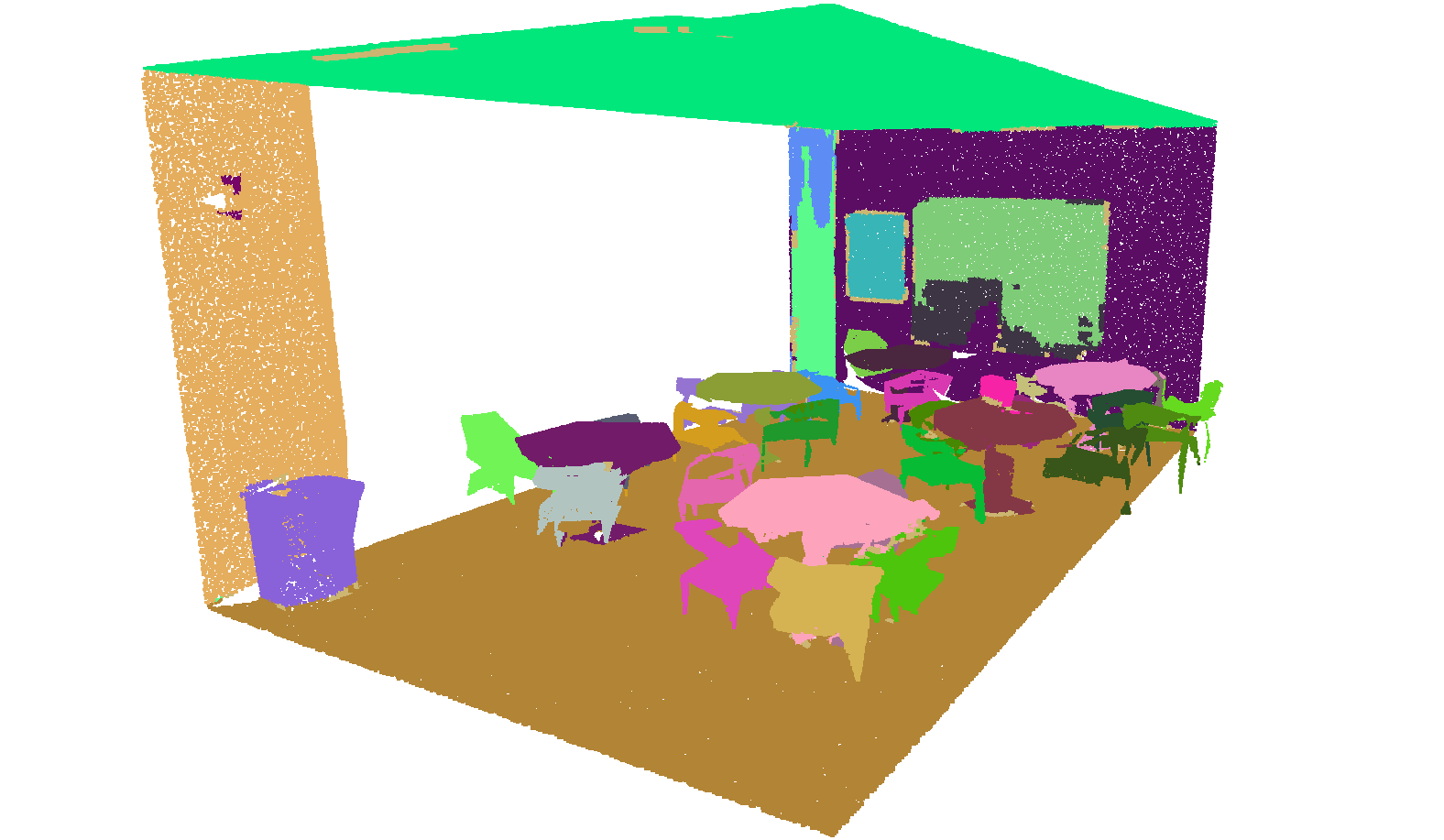} & 
\includegraphics[width=0.19\linewidth]{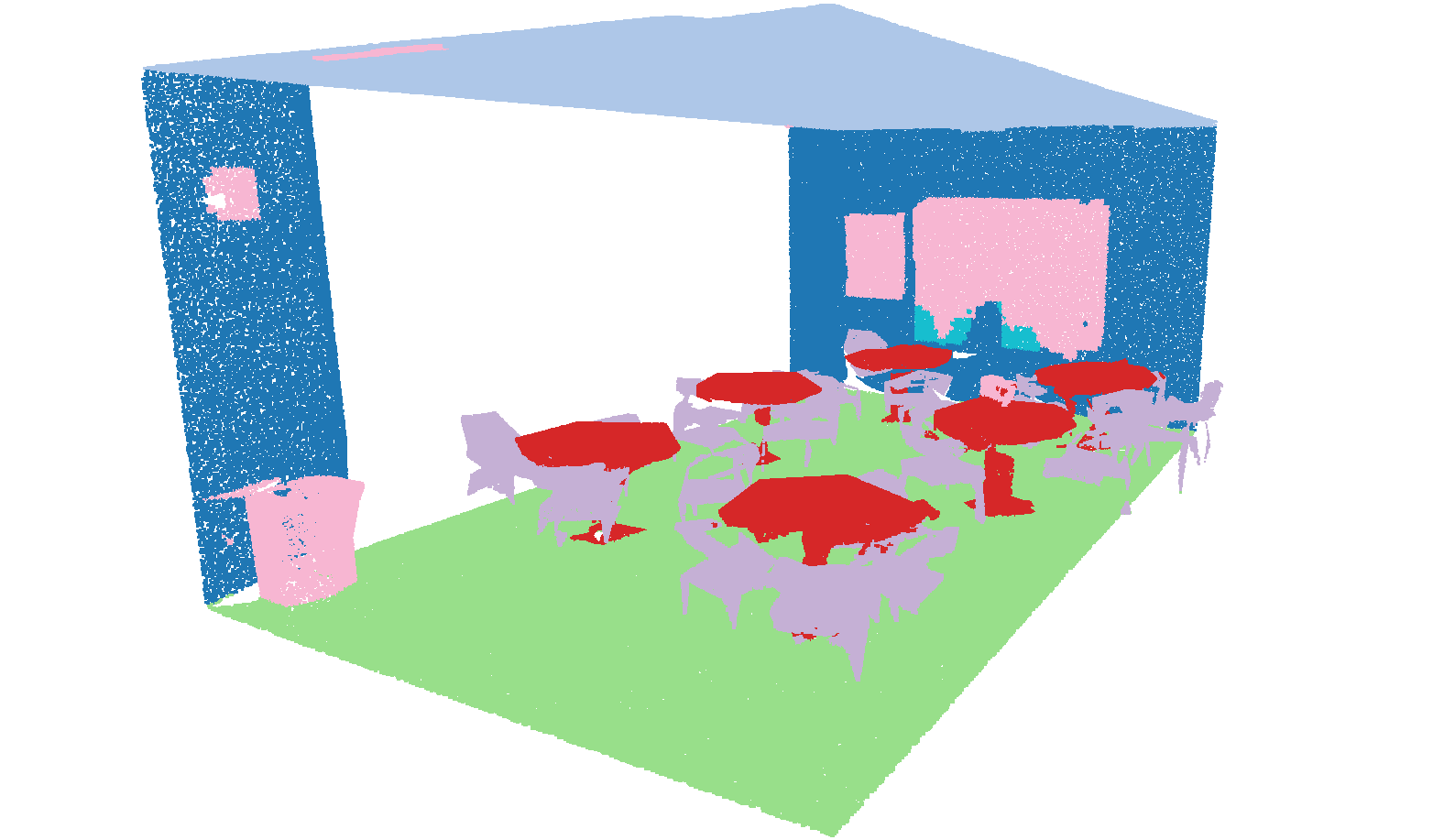} & 
\includegraphics[width=0.19\linewidth]{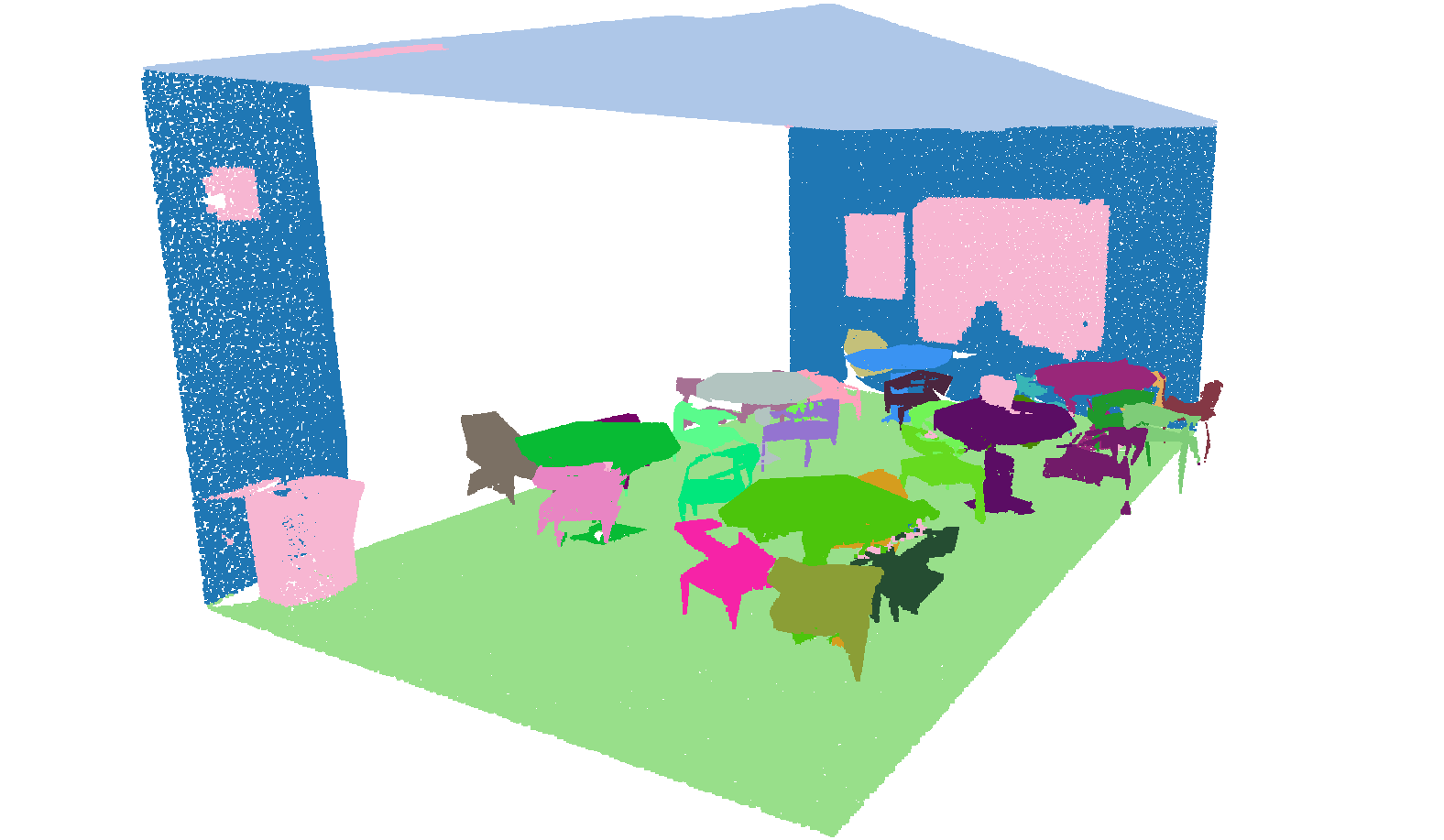} \\
\includegraphics[width=0.19\linewidth]{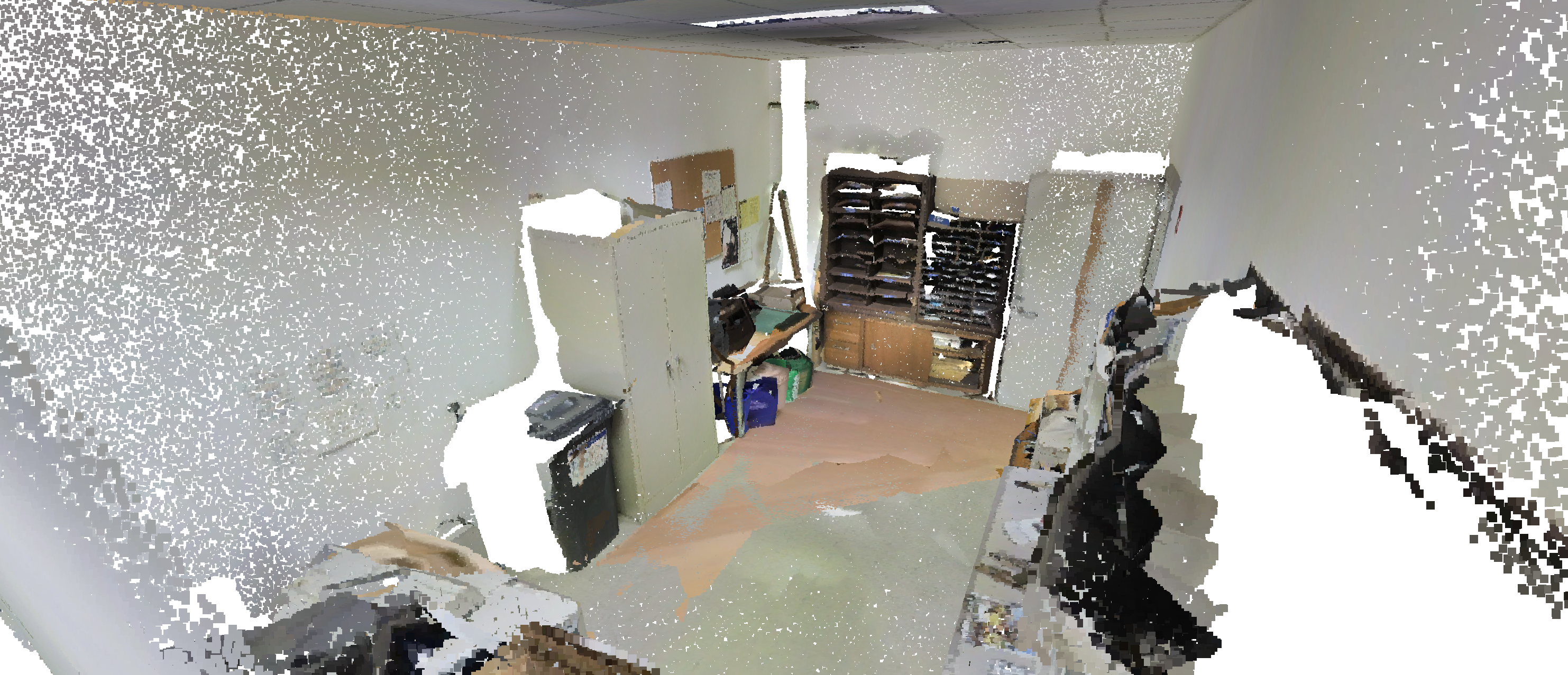} &
\includegraphics[width=0.19\linewidth]{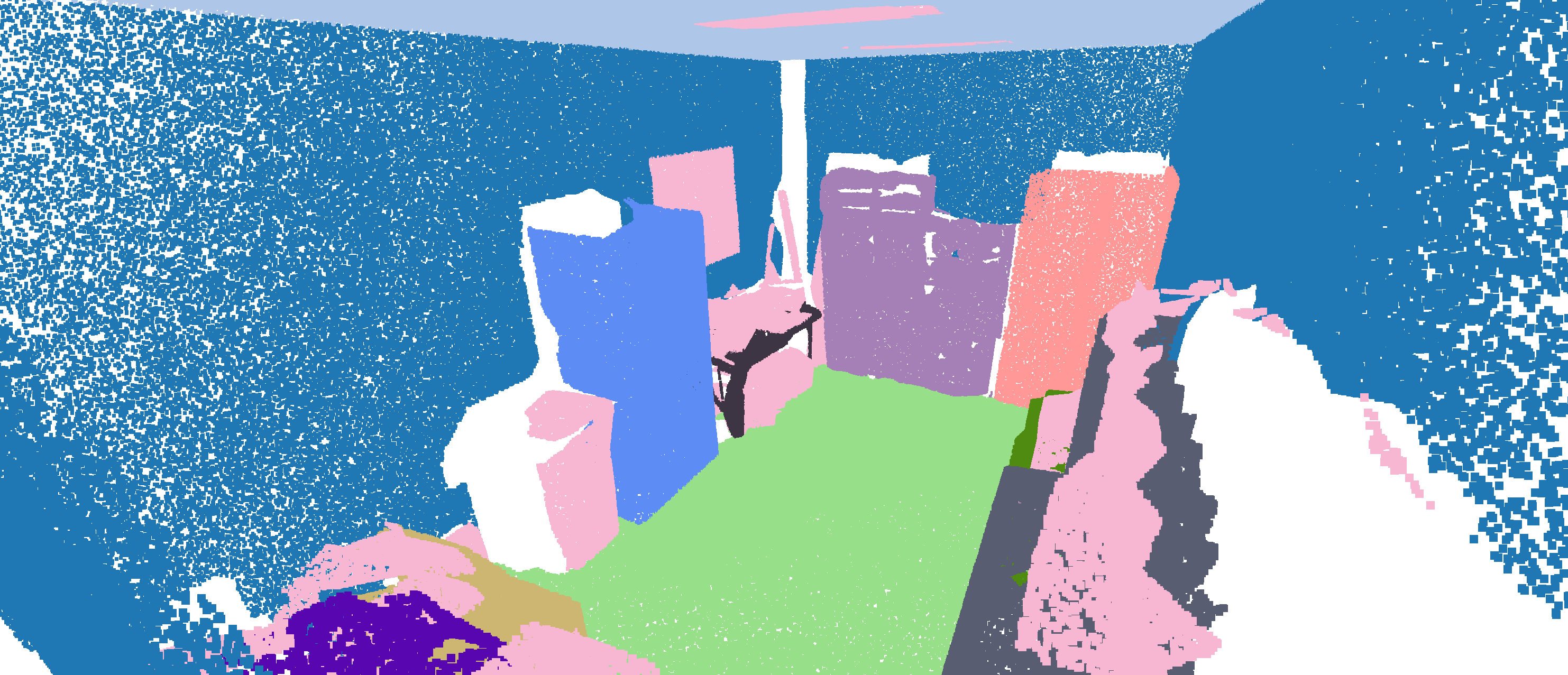} & 
\includegraphics[width=0.19\linewidth]{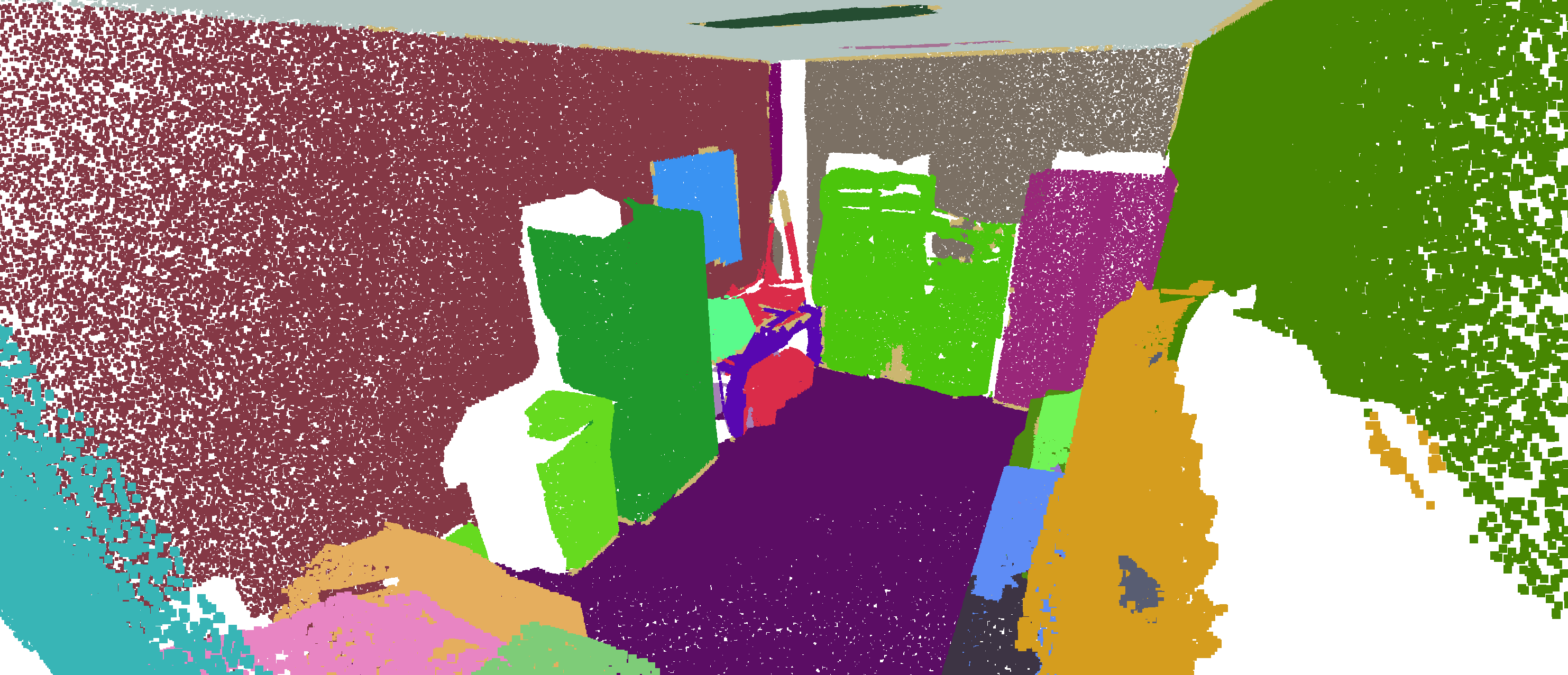} & 
\includegraphics[width=0.19\linewidth]{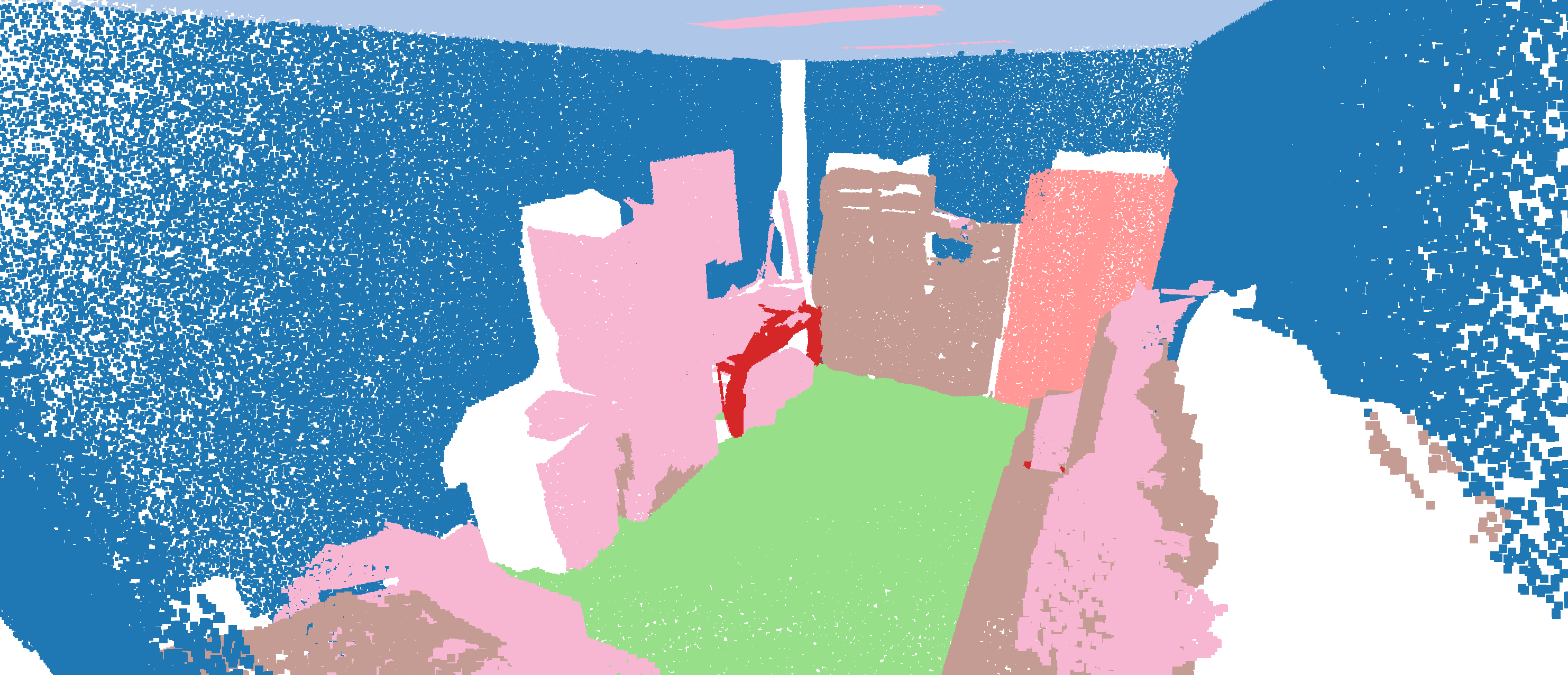} & 
\includegraphics[width=0.19\linewidth]{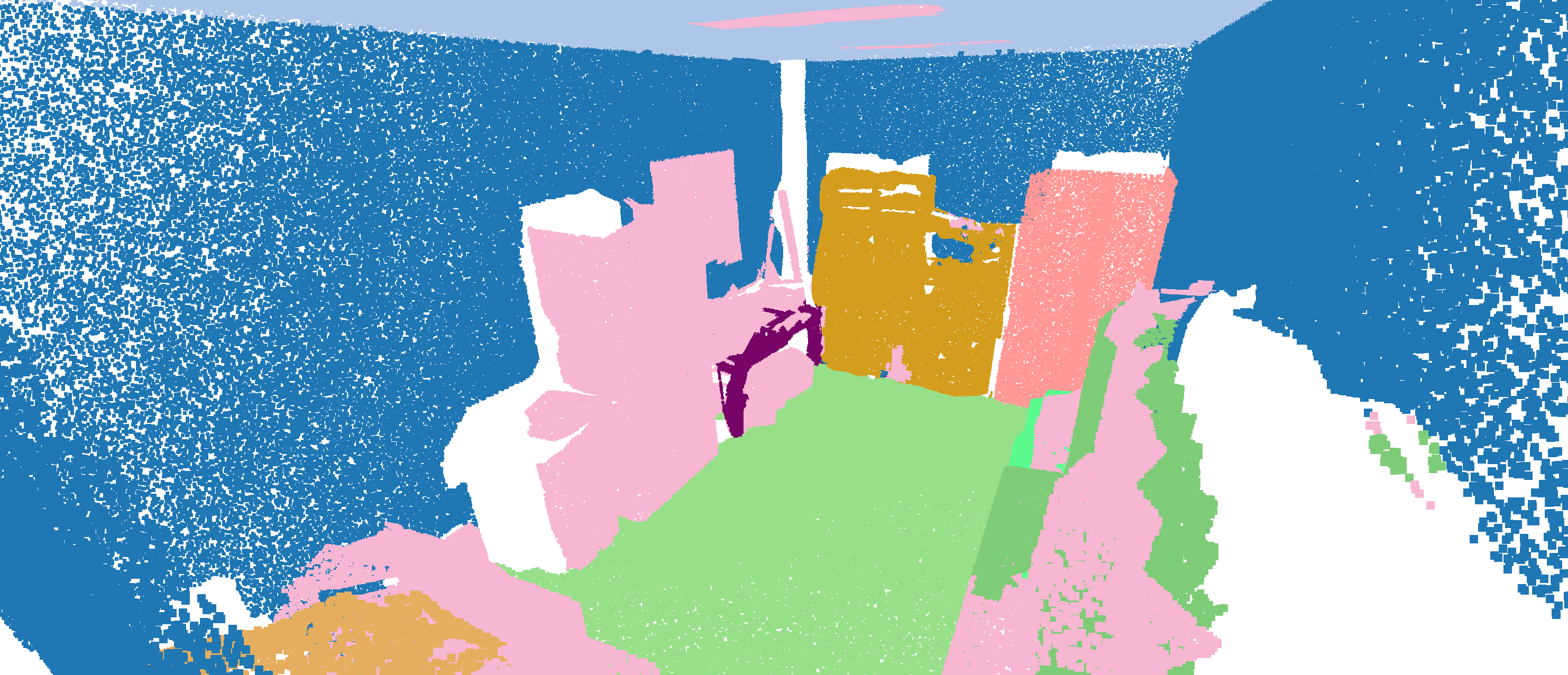} \\
\includegraphics[width=0.19\linewidth]{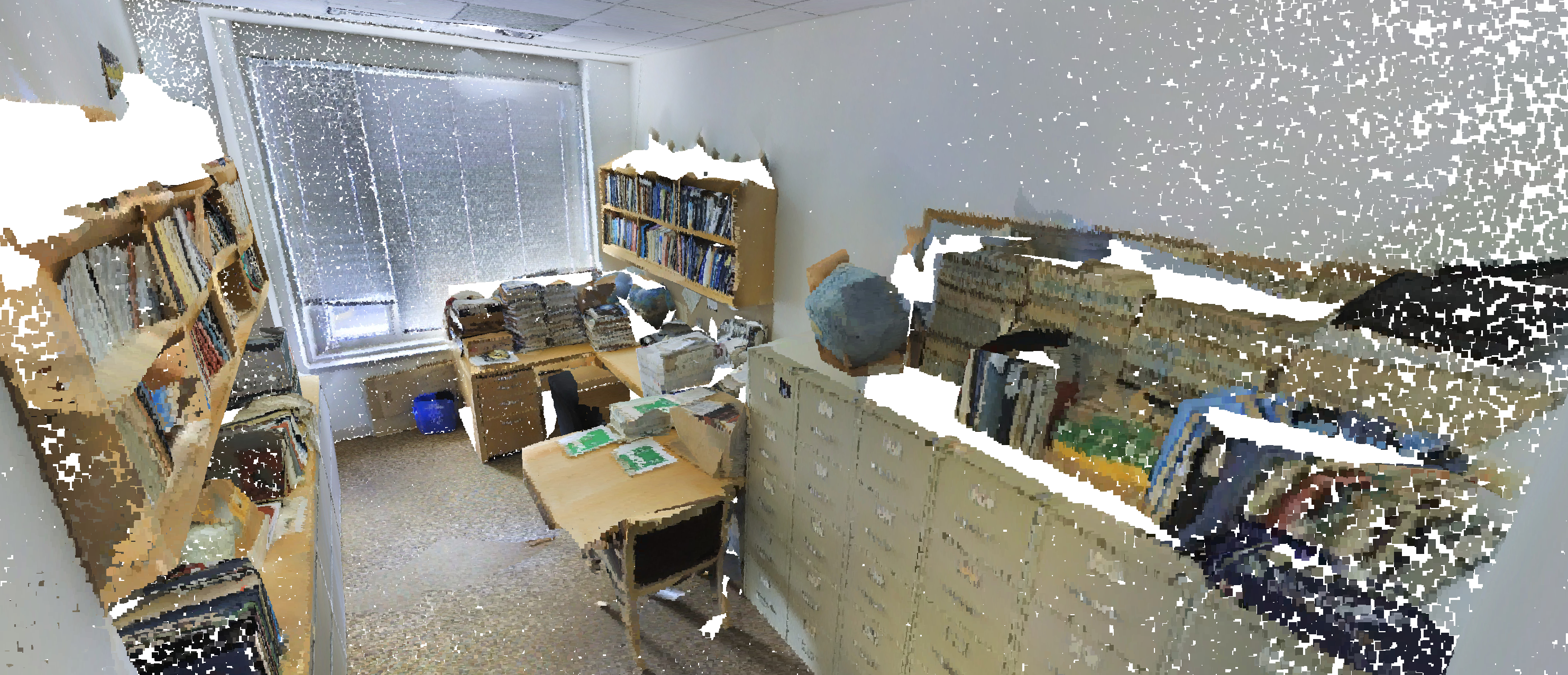} &
\includegraphics[width=0.19\linewidth]{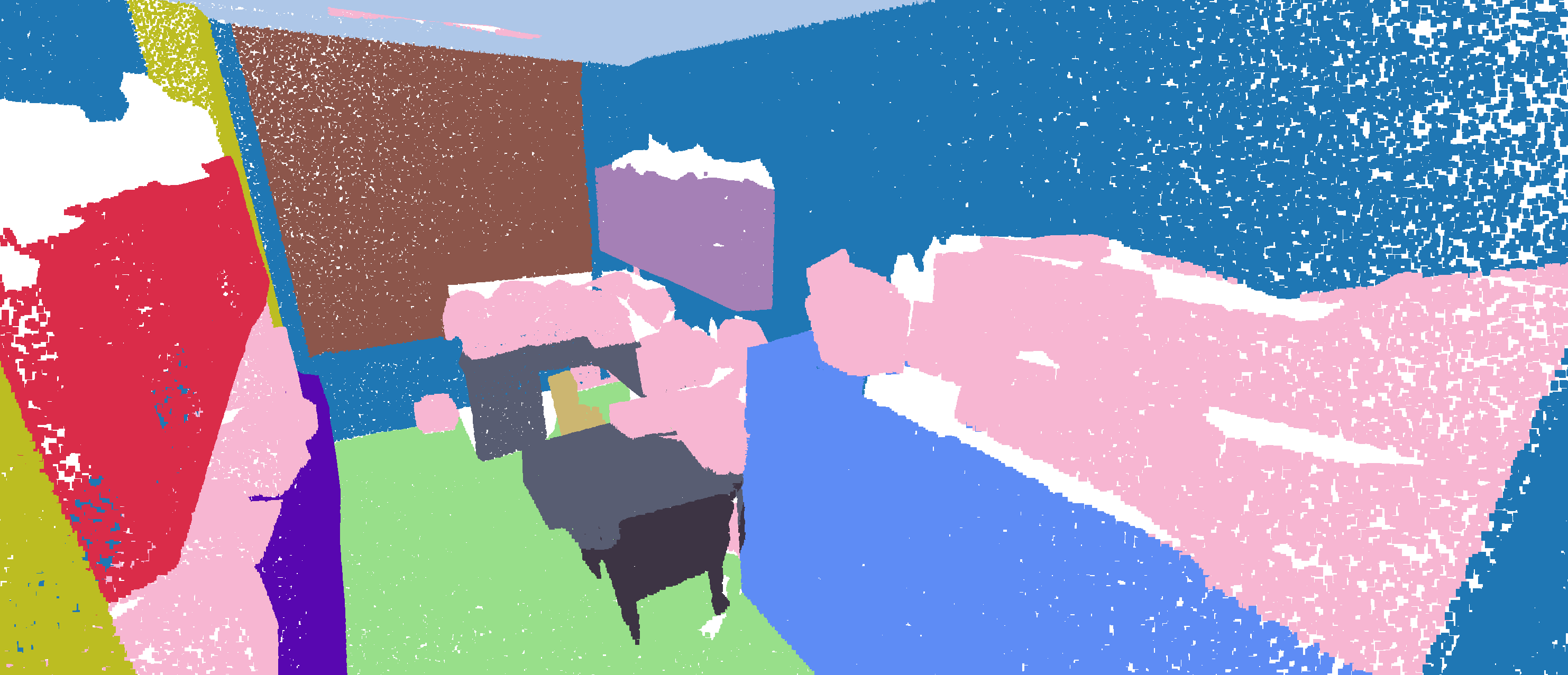} & 
\includegraphics[width=0.19\linewidth]{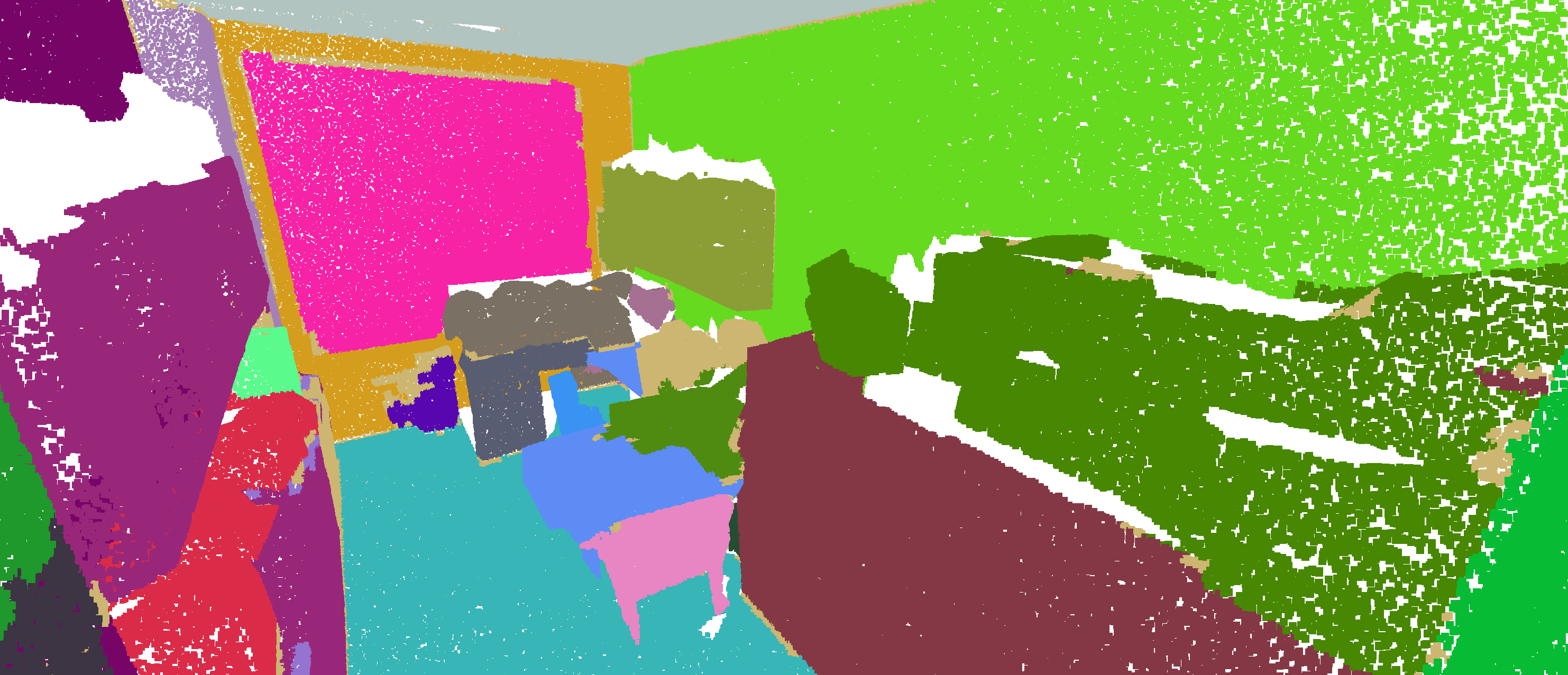} & 
\includegraphics[width=0.19\linewidth]{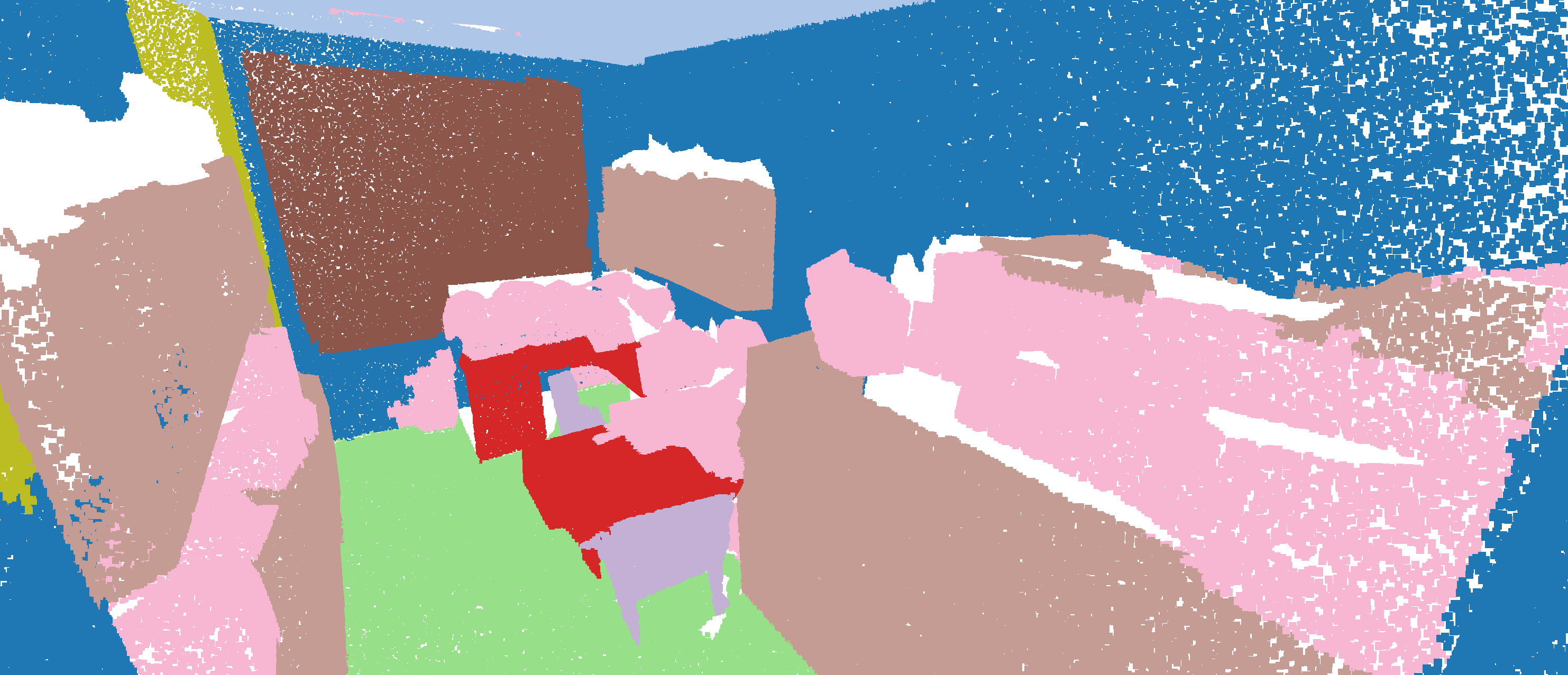} & 
\includegraphics[width=0.19\linewidth]{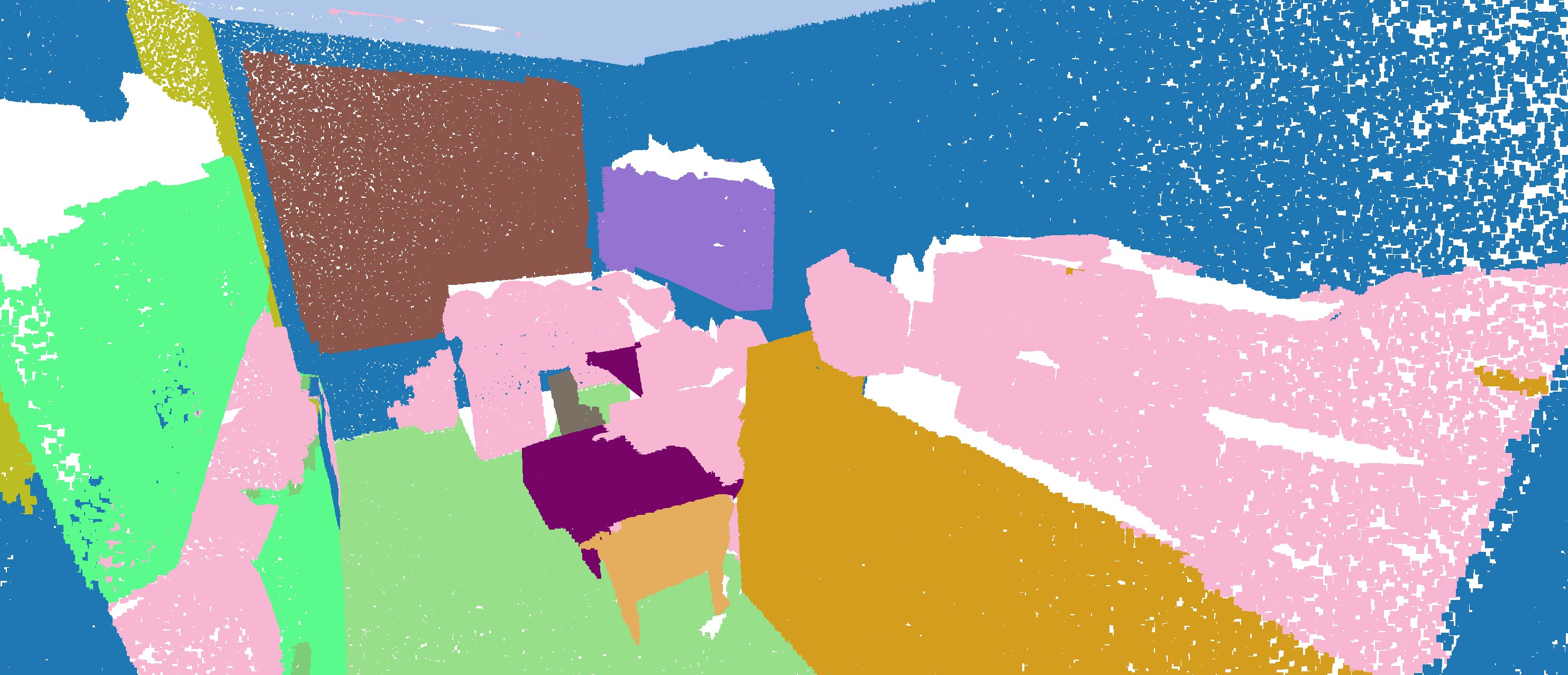} \\
\includegraphics[width=0.19\linewidth]{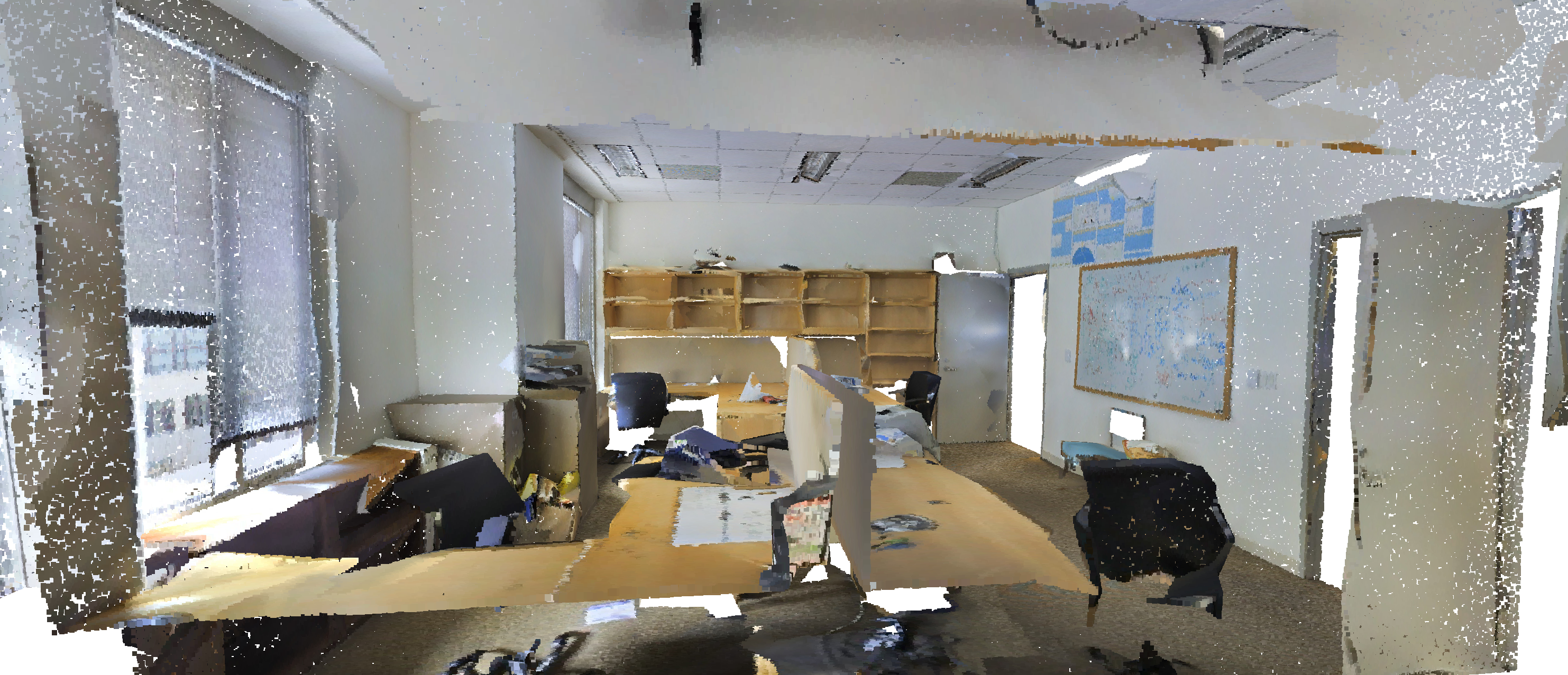} &
\includegraphics[width=0.19\linewidth]{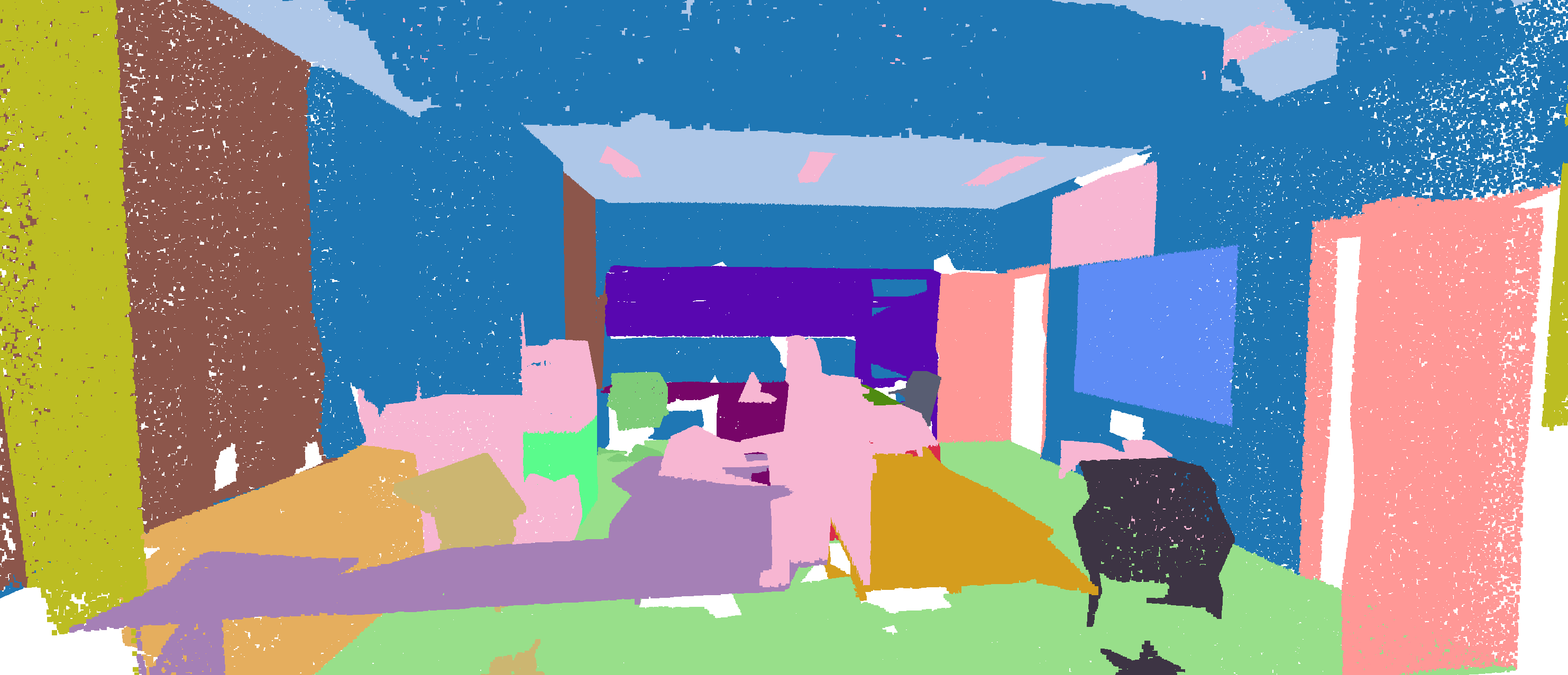} & 
\includegraphics[width=0.19\linewidth]{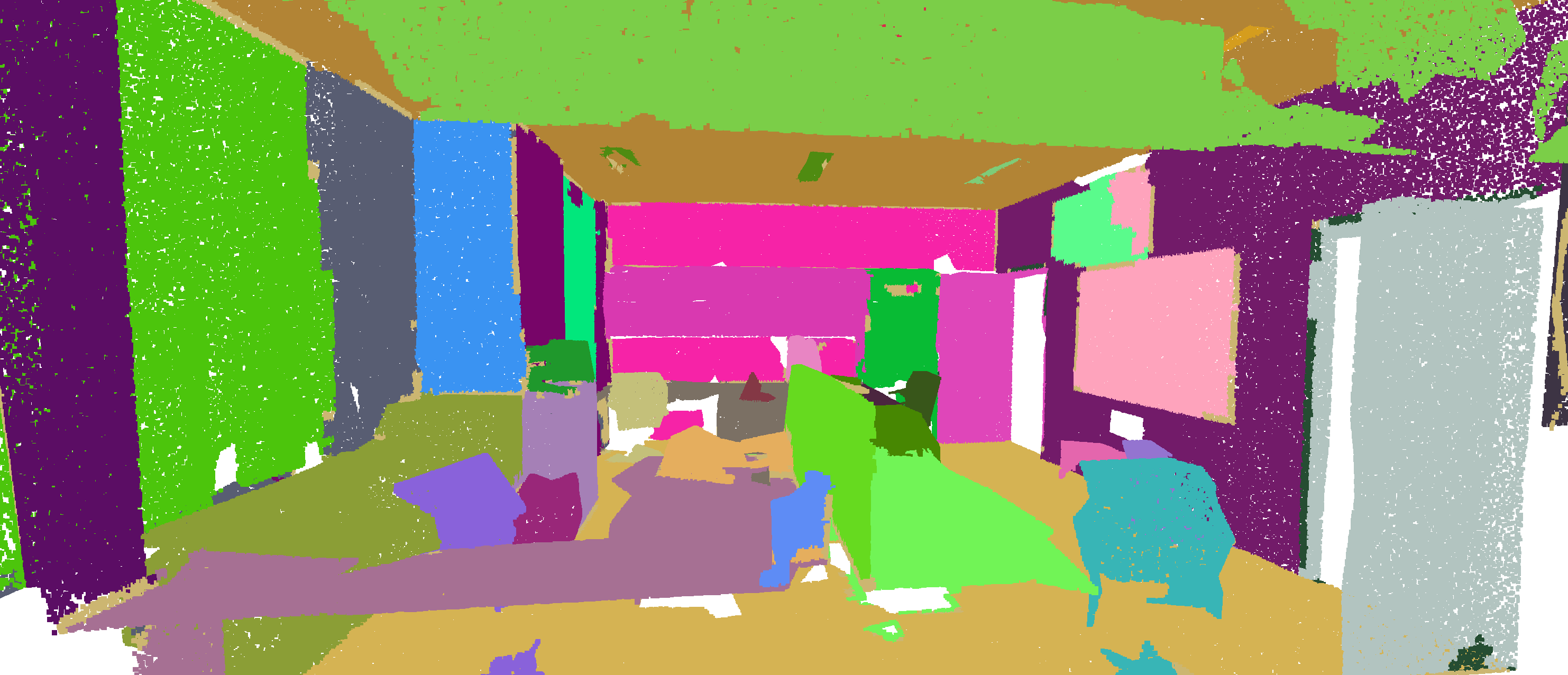} & 
\includegraphics[width=0.19\linewidth]{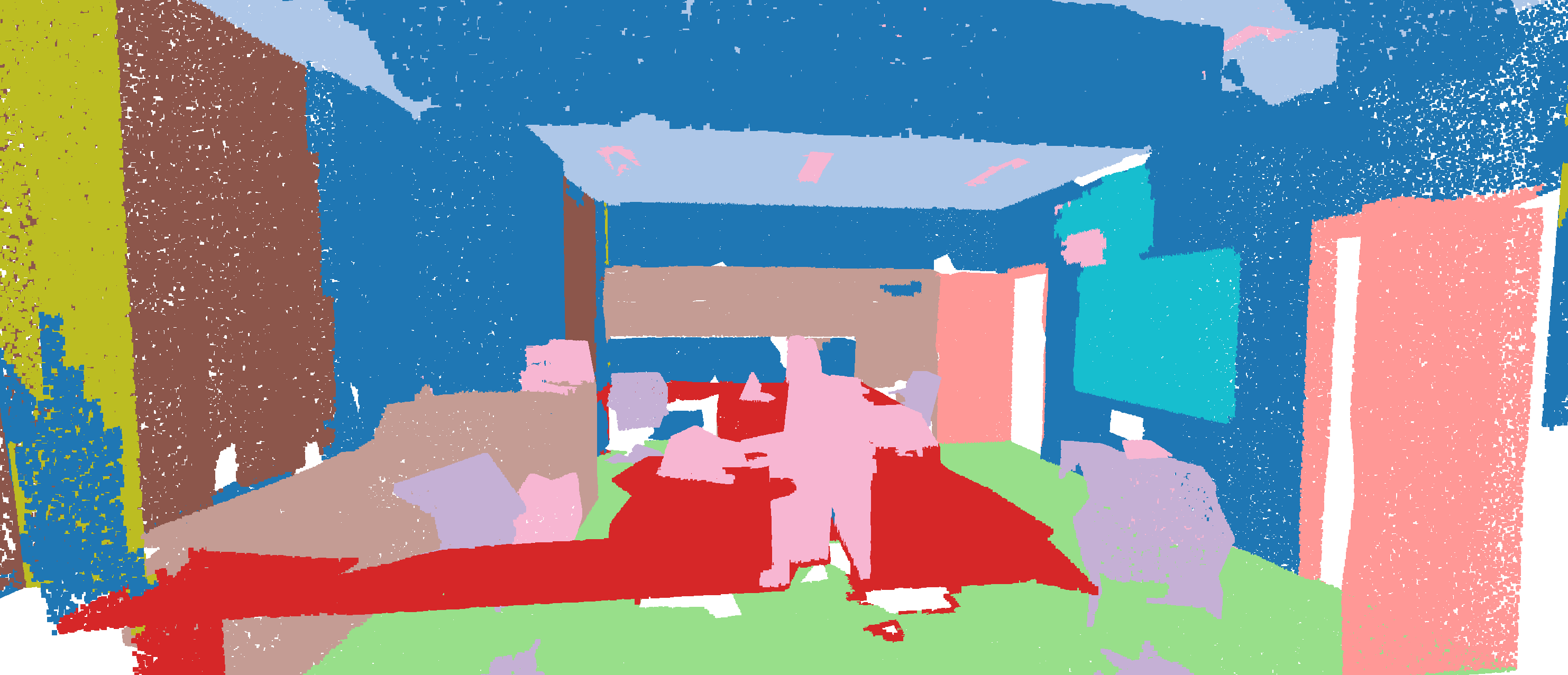} & 
\includegraphics[width=0.19\linewidth]{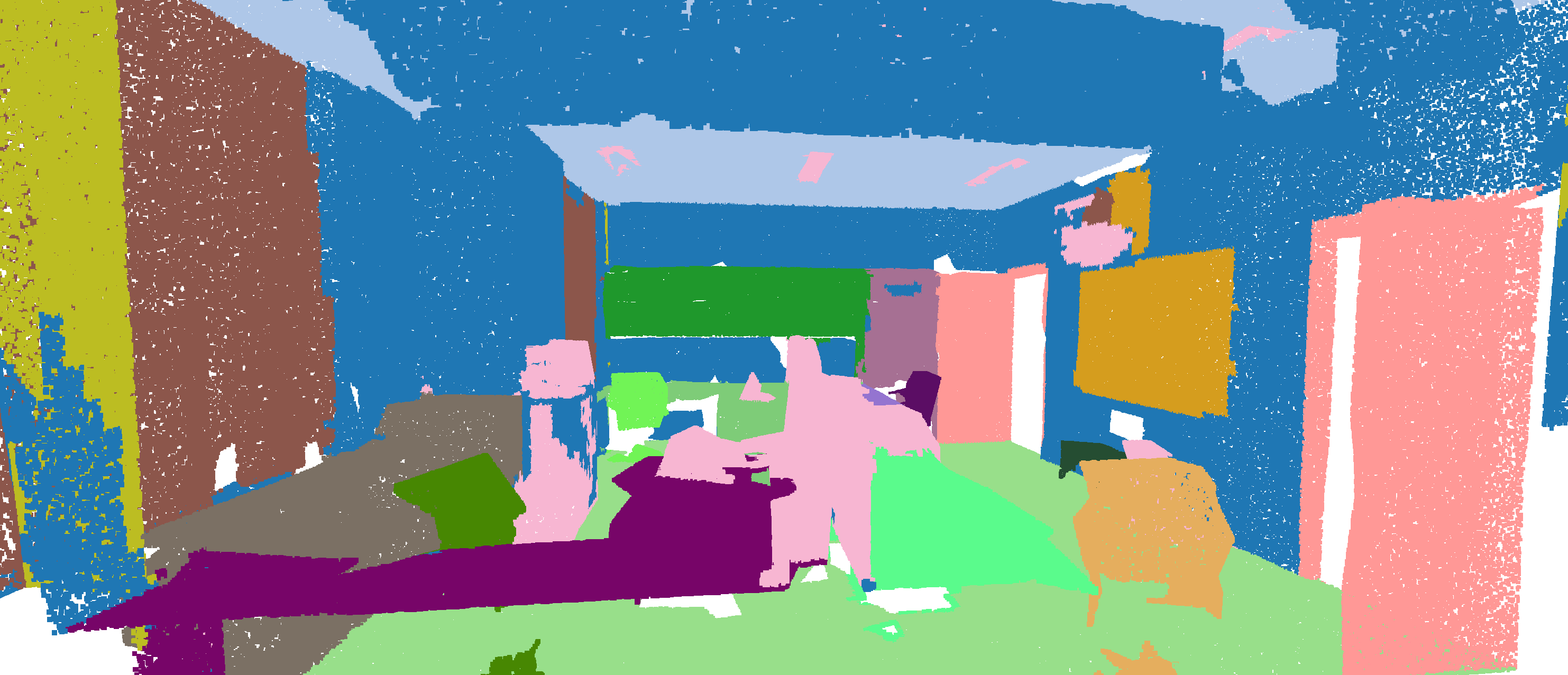} \\
\end{tabular}
\caption{OneFormer3D predictions on the S3DIS Area-5 split. Left to right: an input point cloud, a ground truth panoptic mask, predicted 3D instance, 3D semantic, and 3D panoptic segmentation masks.}
\label{fig:visualition_s3dis}
\end{figure*}

To give an intuition on how the segmentation scores relate to actual segmentation quality, we provide additional visualizations of original and segmented point clouds from the ScanNet (Fig. \ref{fig:visualition_scannet}) and S3DIS (Fig. \ref{fig:visualition_s3dis}) datasets.

\clearpage
{
    \small
    \bibliographystyle{ieeenat_fullname}
    \bibliography{main}
}

\end{document}